\documentclass[review,1p,times]{elsarticle}

\usepackage{multirow}
\usepackage{amssymb}
\usepackage{amsmath}
\usepackage{adjustbox}
\usepackage{amsmath,amssymb,amsfonts}
\usepackage{algorithmic}
\usepackage{graphicx}
\usepackage{textcomp}
\usepackage[table]{xcolor} % for coloring tables
\usepackage{colortbl} % for coloring rows
\colorlet{tableheadcolor}{gray!33} %
\usepackage{minted}

\usepackage{hyperref}
\usepackage{cuted}

\usepackage{subcaption}
\usepackage{array}
\usepackage{booktabs}
\usepackage[ruled,vlined]{algorithm2e}
\usepackage{listings}
\usepackage{multirow}
\usepackage{multicol}

\journal{}

\usepackage[edges]{forest}
\usepackage{tikz}

\def\XS{\xspace}
\def\figureabvr{Figure\XS} 
\def\tableabvr{Table\XS}
\def\eg{\textit{e.g.,}\XS}

\usepackage{bera}% optional: just to have a nice mono-spaced font
\usepackage{listings}
\usepackage{xcolor}

\colorlet{punct}{red!60!black}
\definecolor{background}{HTML}{EEEEEE}
\definecolor{delim}{RGB}{20,105,176}
\colorlet{numb}{magenta!60!black}

\lstdefinelanguage{json}{
    basicstyle=\normalfont\ttfamily,
    numbers=left,
    numberstyle=\scriptsize,
    stepnumber=1,
    numbersep=8pt,
    showstringspaces=false,
    breaklines=true,
    frame=lines,
    backgroundcolor=\color{background},
    literate=
     *{0}{{{\color{numb}0}}}{1}
      {1}{{{\color{numb}1}}}{1}
      {2}{{{\color{numb}2}}}{1}
      {3}{{{\color{numb}3}}}{1}
      {4}{{{\color{numb}4}}}{1}
      {5}{{{\color{numb}5}}}{1}
      {6}{{{\color{numb}6}}}{1}
      {7}{{{\color{numb}7}}}{1}
      {8}{{{\color{numb}8}}}{1}
      {9}{{{\color{numb}9}}}{1}
      {:}{{{\color{punct}{:}}}}{1}
      {,}{{{\color{punct}{,}}}}{1}
      {\{}{{{\color{delim}{\{}}}}{1}
      {\}}{{{\color{delim}{\}}}}}{1}
      {[}{{{\color{delim}{[}}}}{1}
      {]}{{{\color{delim}{]}}}}{1},
}

\definecolor{folderbg}{RGB}{124,166,198}
\definecolor{folderborder}{RGB}{110,144,169}

\def\Size{4pt}
\tikzset{
      folder/.pic={
        \filldraw[draw=folderborder,top color=folderbg!50,bottom color=folderbg]
          (-1.05*\Size,0.2\Size+5pt) rectangle ++(.75*\Size,-0.2\Size-5pt);  
        \filldraw[draw=folderborder,top color=folderbg!50,bottom color=folderbg]
          (-1.15*\Size,-\Size) rectangle (1.15*\Size,\Size);
      }
      file/.pic={
    \filldraw[draw=black!50, top color=white, bottom color=black!10]
      (-0.5*\Size,-0.5*\Size) rectangle ++(\Size,\Size);
  }
    }

\begin{document}

\begin{frontmatter}

\title{Teeth3DS+: An Extended Benchmark for Intraoral 3D Scans Analysis}
\author[label1,label2]{Achraf Ben-Hamadou\corref{cor1}}\ead{achraf.benhamadou@crns.rnrt.tn}
\author[label1,label2]{Nour Neifar}\ead{}
\author[label1,label2]{Ahmed Rekik}\ead{}
\author[udini]{Oussama Smaoui}\ead{}
\author[udini]{Firas Bouzguenda}\ead{}
\author[inria]{Sergi Pujades}\ead{}
\author[inria]{Edmond Boyer}\ead{}
\author[udini]{Edouard Ladroit}\ead{}

\address[label1]{SMARTS Laboratory, Technopark of Sfax, Sakiet Ezzit 3021, Sfax, Tunisia}
\address[label2]{Digital Research Center of Sfax, Technopark of Sfax, Sakiet Ezzit 3021, Sfax, Tunisia}

\address[udini]{Udini, 37 BD Aristide Briand, 13100 Aix-En-Provence, France}
\address[inria]{Inria, Univ. Grenoble Alpes, CNRS, Grenoble INP, LJK, France}

\cortext[cor1]{Corresponding author}

%% Abstract
\begin{abstract}
Intraoral 3D scanning is now widely adopted in modern dentistry and plays a central role in supporting key tasks such as tooth segmentation, detection, labeling, and dental landmark identification. Accurate analysis of these scans is essential for orthodontic and restorative treatment planning, as it enables automated workflows and minimizes the need for manual intervention. However, the development of robust learning-based solutions remains challenging due to the limited availability of high-quality public datasets and standardized benchmarks. This article presents Teeth3DS+, an extended public benchmark dedicated to intraoral 3D scan analysis. Developed in the context of the MICCAI 3DTeethSeg and 3DTeethLand challenges, Teeth3DS+ supports multiple fundamental tasks, including tooth detection, segmentation, labeling, 3D modeling, and dental landmark identification. The dataset consists of rigorously curated intraoral scans acquired using state-of-the-art scanners and validated by experienced orthodontists and dental surgeons. In addition to the data, Teeth3DS+ provides standardized data splits and evaluation protocols to enable fair and reproducible comparison of methods, with the goal of fostering progress in learning-based analysis of 3D dental scans. 
\end{abstract}

\begin{keyword}
Teeth3DS \sep Teeth3DS+ \sep intraoral 3D scans \sep 3D point cloud \sep 3D segmentation \sep dentistry
\end{keyword}

\end{frontmatter}
\section{Introduction}
 \label{sec:introduction}
The clinical demands of dentistry, from diagnostic assessment to restorative and orthodontic treatment, continue to expand, requiring ever higher levels of precision and consistency. As these applications have grown in complexity, digital technologies have become increasingly central to delivering precise and predictable outcomes. In particular, advanced intraoral scanners (IOSs) are now widely adopted in orthodontics, providing high-resolution three-dimensional representations of the dentition \cite{Farhat_2025_BMVC,eggmann2024recent}. These technologies capture detailed anatomical information and generate accurate virtual models, enabling clinicians to evaluate multiple treatment strategies, such as tooth extraction planning, realignment, and smile design, with considerably improved accuracy \cite{krenmayr2025evaluating}. In addition, the use of digital models significantly reduces traditionally manual and repetitive tasks,  optimizing workflows and saving clinical time \cite{siqueira2021intraoral}. Building on these digital representations, recent work has focused on automated teeth segmentation, labeling, and landmark identification tasks that are essential for efficient orthodontic analysis and planning.

However, accurate teeth segmentation, labeling, and landmark identification in IOS data remain challenging due to substantial inter-class variations (\eg., similar tooth shapes, ambiguous jaw boundaries) and intra-class variations (e.g., damaged teeth, the presence of braces). Additional difficulties arise from the close boundaries between teeth and gingiva, as well as abnormalities such as crowding, misalignment, or missing teeth. The landmark identification task is further complicated by the complex geometry of individual teeth, variations in shape, size, and orientation, and inter-individual differences in dental arch forms, alignment, and occlusal relationships. To address these challenges, the development of advanced techniques has become a major research focus, particularly with the rise of deep learning. However, the number of studies addressing teeth segmentation and labeling \cite{MeshSNet2019,xu20183d,sun2020automatic,zhao_3d_2021,ma_efficient_2022,zhao_two-stream_2022,rekik2025tseglab}, and especially dental landmark identification, remains limited. Traditional segmentation and labeling approaches often rely on geometric features such as curvature thresholding \cite{kumar2011improved} or active contours \cite{yaqi2010computer}, and frequently require manual intervention. Recent advances in 3D shape analysis have explored deep-learning-based solutions. Early methods focused on extracting relevant features using CNNs applied to 2D image representations derived from 3D scans. More recently, researchers have applied deep learning architectures directly to 3D dental meshes \cite{8984309} to detect and segment teeth in IOS data. Although these approaches can be faster and more automated, they still face limitations related to point-cloud irregularities, downsampling, and 3D pose variations. Studies on landmark identification can similarly be divided into two main categories: traditional methods \cite{woodsend2021automatic,triarjo2023automatic} and deep-learning-based approaches \cite{lang2022dentalpointnet,lang2021dllnet}. Some of these methods adopt a unified pipeline, performing tooth segmentation followed by landmark identification.

A major challenge in this domain is the absence of publicly available datasets and standardized benchmarks, which limits consistent evaluation, fair comparison of methods, and the development of more accurate and robust algorithms. To bridge this gap, we introduce Teeth3DS+, the first comprehensive public benchmark of 3D intraoral scans. The dataset was specifically created to support the MICCAI challenges 3DTeethSeg \cite{ben20233dteethseg} and 3DTeethLand \cite{benhamadou2025detectingdentallandmarksintraoral}, covering a wide range of tasks including tooth detection, segmentation, labeling, and landmark localization. By offering a unified benchmark, our dataset aims to accelerate research and foster a strong research community dedicated to advancing 3D dental analysis.

The main contributions of this paper are as follows:
\begin{itemize}
 \item We introduce Teeth3DS+, an extended public benchmark for 3D intraoral scans, validated for multiple tasks including tooth detection, segmentation, labeling, and landmark identification.

 \item We provide a standardized evaluation protocol, including data splits and metrics, enabling fair and reproducible comparison of learning-based methods.

 \item We release an open and extensible implementation integrated within the PyTorch Geometric framework to encourage research on 3D dental shape analysis.

\end{itemize}

\section{Method}\label{sec:methods}
The process for creating the dataset consists of several steps, including data collection, annotation, preprocessing, and preparation of data records. Each step was carefully designed to ensure accuracy and consistency.
\subsection{Patients and intraoral scans collection}

The database was initially developed for the 3DTeethSeg challenge to support research on intraoral 3D scan analysis, including tooth detection, segmentation, and labeling. To ensure compliance with the European General Data Protection Regulation (GDPR), all data were collected under strict privacy protocols, and patient information was fully anonymized. Experienced orthodontists and dental surgeons from partner clinics in France and Belgium participated in the acquisition process. 

Scans were obtained using three intraoral scanners: Primescan (Dentsply), Trios3 (3Shape), and iTero Element 2 Plus. These devices produced 3D reconstructions with an accuracy of 10–90 micrometers and a point density of 30–80 pts/mm². No additional equipment was used during acquisition. Each patient contributed two scans, corresponding to the upper and lower jaws, resulting in a total of 1800 scans from 900 patients. Clinical cases included 50\% orthodontic and 50\% prosthetic treatments. The patient population consisted of 50\% male and 50\% female, with approximately 70\% under 16 years, 27\% between 16 and 59 years, and 3\% over 60 years.
\begin{figure*}[t]
    \centering
    \includegraphics[draft=false,width=\textwidth]{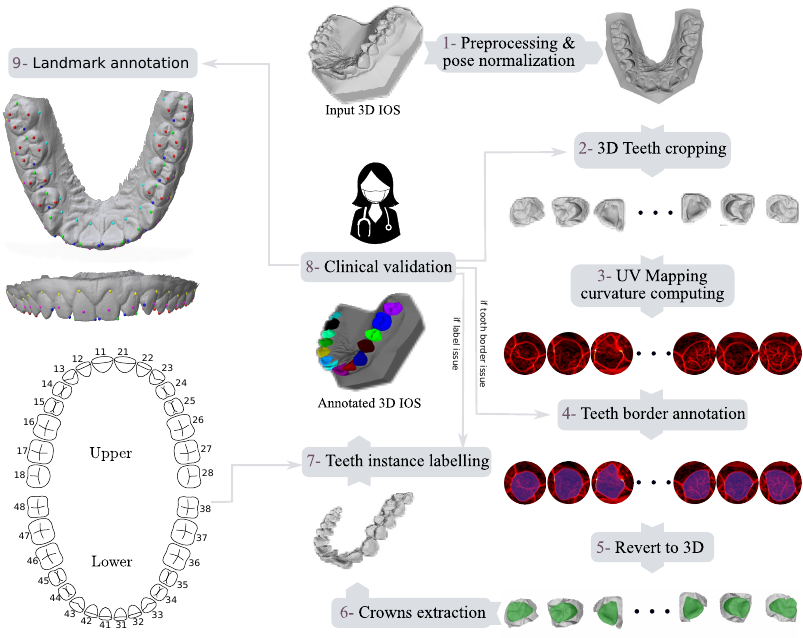}
    \caption{Illustration of our annotation process. An input 3D intraoral scan is annotated following nine steps, starting with preprocessing and pose normalization and ending with clinical validation. The workflow includes teeth detection, segmentation, labeling, and dental landmark annotation. During the validation stage, the clinical expert may return the annotation to steps 2, 4, 7, or 9 to address, respectively, missing teeth, incorrect tooth boundaries, erroneous tooth instance labeling, or landmark annotation errors.
    }   
    \label{fig:annotation_process}
\end{figure*}

To extend the dataset with an additional task, a subset of 340 scans from the original collection was annotated with dental landmark identification as part of the MICCAI 3DTeethLand challenge. All scans enriched with landmark annotations are included within the segmentation dataset, ensuring full consistency across tasks and providing a focused benchmark for evaluating landmark identification methods. 

\subsection{Data annotation and preprocessing}
The data annotation and preprocessing pipeline was designed to generate high-quality, expert-validated ground truth for tooth segmentation, labeling, and dental landmark identification from intraoral 3D scans. Annotation was performed in collaboration with clinical experts with over ten years of experience in orthodontics, dental surgery, and endodontics. The workflow is illustrated in \figureabvr~\ref{fig:annotation_process} and involves nine sequential steps.

Initially, the 3D scans undergo preprocessing (steps 1 and 2 in \figureabvr~\ref{fig:annotation_process}), which includes the removal of degenerate or redundant mesh faces and duplicated or irrelevant vertices. The dental meshes are then automatically centered and aligned with the occlusal plane using principal component analysis, improving tooth visibility and normalizing the 3D pose of all intraoral scans. 

Next, an internal annotation tool is used to manually crop each tooth within a tight spherical region that encompasses the target tooth along with neighboring teeth and gingiva. Step 3 applies UV mapping to the cropped meshes, flattening them to display the maximum 3D curvature, which facilitates the annotation of tooth boundaries. This transformation is achieved via harmonic parameterization, a fixed-boundary algorithm that computes 2D coordinates of the flattened mesh vertices using two harmonic functions \cite{eck1995multiresolution}. Vertices are first mapped to a circle and the remaining coordinates are calculated under harmonic constraints. This approach allows annotators to draw 2D polygons along tooth boundaries without altering the 3D perspective, while the 3D curvature overlay highlights boundary details.  

Following manual annotation of the UV maps (step 4), the defined tooth boundaries are back-propagated to the original 3D crops in step 5, resulting in individually segmented teeth within a common 3D coordinate system. Step 6 prepares all crowns of teeth for manual labeling, performed in step 7 according to the FDI World Dental Federation numbering system \cite{keiser1971federation}. Step 8 consists of visual inspection and validation of the annotated 3D scans by clinical partners. This review targets potential errors such as missing tooth annotations (return to step 2), incorrect boundary annotations (return to step 4), or mislabeled teeth (return to step 7). The validation and correction cycle is iteratively repeated until all scans are fully segmented, labeled, and clinically verified.

%Landmarks part 
The dental landmark annotation process was carried out iteratively by trained annotators using the same internal tool (step 9). Annotators were instructed to identify and label every tooth landmark class on each tooth within the intraoral 3D scans. Following this, a final validation phase was carried out. Each scan was initially annotated by a single annotator, during which all annotations were systematically reviewed and verified by the three clinical evaluators of the challenge. This multi-step procedure ensures both accuracy and reliability of the resulting landmark annotations. 
These landmarks serve as essential reference points for orthodontic and restorative procedures, enabling precise assessment of tooth orientation, spacing, and occlusal relationships, and supporting detailed diagnostic and treatment planning. A specific set of dental landmarks for each tooth is defined to facilitate precise analysis of tooth position and alignment: 

\begin{enumerate}
\item \textit{Mesial} and \textit{Distal} landmarks (blue and green points, respectively) are positioned on the proximal surfaces of the tooth and are essential for assessing tooth alignment within the arch. The mesial point lies closer to the midline of the dental arch, while the distal point is farther from it. 

\item \textit{Cusp} landmarks (red points) correspond to the highest elevations on the occlusal (chewing) surface. They are crucial for evaluating occlusion relationships between the upper and lower teeth. Each cusp point is defined as the highest point of a cusp.
 
\item \textit{Inner} and \textit{Outer} landmarks (cyan and yellow points, respectively) are essential for determining the spatial orientation of teeth. They are located at the boundaries of the tooth, where it meets the gingiva. The inner point faces the tongue or palate, whereas the outer point faces the cheeks or lips. 

\item \textit{Facial} landmarks (magenta points) are located at the center of the facial surfaces of teeth, visible from the front of the mouth. They are also important for assessing tooth spatial orientation.

\end{enumerate}

\section{Data records}\label{sec:ResultingDataset}

The overall file structure is depicted in \figureabvr \ref{fig:data_structure}. Data records are grouped by patient. All 3D intraoral scans are provided in OBJ format. Tooth instance and class annotations (labels) corresponding to each vertex in the OBJ file are provided in JSON format. An example of such a JSON file is shown below:

\vspace*{1cm}
\includegraphics[scale=1.3]{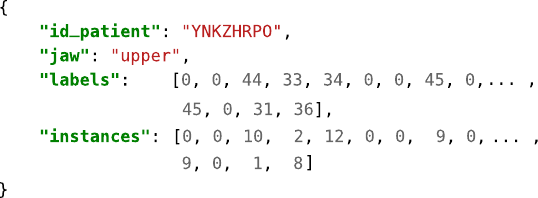}

\vspace*{1cm}
\noindent Dental landmark annotations are provided in JSON format. An example of the landmarks file format is shown below:

\vspace*{1cm}
\includegraphics[scale=1.3]{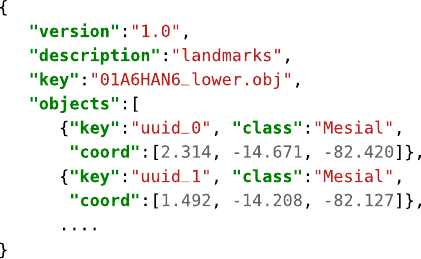}

\def\verticalsep{1.3em} 
\begin{figure*}[t]
\centering
\begin{minipage}[t]{0.49\textwidth}
\begin{forest}
  for tree={
    l sep=0.5em,
    s sep=0.5em,
    font=\ttfamily,
    grow'=0,
    child anchor=west,
    parent anchor=south,
    anchor=west,
    calign=first,
    inner xsep=7pt,
    edge path={
      \noexpand\path [draw, \forestoption{edge}]
      (!u.south west) +(7.5pt,0) |- (.child anchor) pic {folder} \forestoption{edge label};
    },
    file/.style={
      edge path={
        \noexpand\path [draw, \forestoption{edge}]
        (!u.south west) +(7.5pt,0) |- (.child anchor) \forestoption{edge label};
      },
      inner xsep=2pt,
      font=\small\ttfamily,
      s sep=0.05em
    },
    before typesetting nodes={
      if n=1 {insert before={[,phantom]}} {}
    },
    fit=band,
    before computing xy={l=15pt},
    l sep=1.5em, s sep=\verticalsep
  }
  [Training
    [Upper
      [01MCM1UK
        [01MCM1UK\_upper\_\_kpt.json, file]
        [01MCM1UK\_upper.obj, file]
        [01MCM1UK\_upper.json, file]
      ]
      [2TPJ3FG9
        [\dots, file]
      ]
      [\dots, file]
    ]
    [Lower
      [QF3P9YRI
        [QF3P9YRI\_lower\_\_kpt.json, file]
        [QF3P9YRI\_lower.obj, file]
        [QF3P9YRI\_lower.json, file]
      ]
      [X273URFW
        [\dots, file]
      ]
      [\dots, file]
    ]
  ]
\end{forest}
\end{minipage}%
\hfill
\begin{minipage}[t]{0.49\textwidth}
\begin{forest}
  for tree={
    l sep=0.5em,
    s sep=0.5em,
    font=\ttfamily,
    grow'=0,
    child anchor=west,
    parent anchor=south,
    anchor=west,
    calign=first,
    inner xsep=7pt,
    edge path={
      \noexpand\path [draw, \forestoption{edge}]
      (!u.south west) +(7.5pt,0) |- (.child anchor) pic {folder} \forestoption{edge label};
    },
    file/.style={
      edge path={
        \noexpand\path [draw, \forestoption{edge}]
        (!u.south west) +(7.5pt,0) |- (.child anchor) \forestoption{edge label};
      },
      inner xsep=2pt,
      font=\small\ttfamily,
      s sep=0.05em
    },
    before typesetting nodes={
      if n=1 {insert before={[,phantom]}} {}
    },
    fit=band,
    before computing xy={l=15pt},
    l sep=1.5em, s sep=\verticalsep
  }
  [Testing
    [Upper
      [P0LH5JC8
       [P0LH5JC8\_upper\_\_kpt.json, file]
        [P0LH5JC8\_upper.obj, file]
        [P0LH5JC8\_upper.json, file]
      ]
      [ECZA08SG
        [\dots, file]
      ]
      [\dots, file]
    ]
    [Lower
      [FFET1F1B
       [FFET1F1B\_lower\_\_kpt.json, file]
        [FFET1F1B\_lower.obj, file]
        [FFET1F1B\_lower.json, file]
      ]
      [8XVU7PN9
        [\dots, file]
      ]
      [\dots, file]
    ]
  ]
\end{forest}
\end{minipage}
\caption{Structure of the data records. Training and testing data are separated. All data are structured by jaws and patient UUID. OBJ files represent the raw intraoral scans. JSON files provide the annotation corresponding to the OBJ files. Landmarks annotations are provided in JSON files named with the suffix kpt, which stands for keypoints. }
\label{fig:data_structure}
\end{figure*}

\clearpage
\section{Technical Validation}
\begin{figure*}[h]
\centering
\begin{tabular}{cc}
  \includegraphics[draft=false,width=0.5\textwidth]{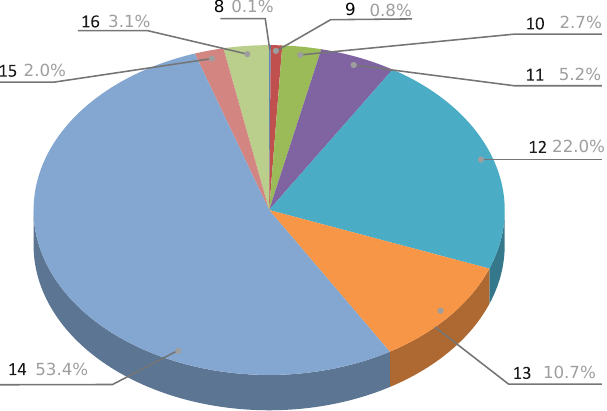} & \includegraphics[draft=false,width=0.5\textwidth]{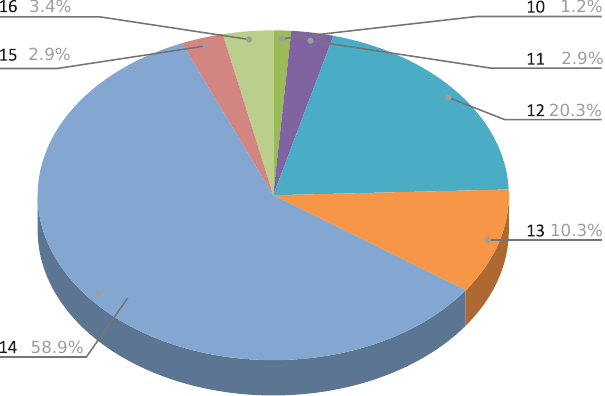} \\
(a) & (b) 
\end{tabular}
\caption{Upper (a) and lower (b) statistics regarding the number of teeth per jaw.}\label{fig:upper_statistics}
\end{figure*}

Dataset statistics regarding the tooth detection, segmentation, and labeling are provided in Table~\ref{tab:description}. We selected two-thirds of the scans for training and the rest for testing, totaling 1200 and 600 scans, respectively. The scan selection was performed in such a way that the same training/testing proportion is maintained at the level of tooth labels as well, ensuring balanced data. For example, there are 591 teeth in the training subset out of a total of 888 teeth with label 12, which corresponds to approximately $66.55\%$. Additionally, \figureabvr~\ref{fig:upper_statistics} shows the distribution of the number of teeth per jaw, indicating that jaws with 14 teeth are the most represented cases in the dataset.

\definecolor{tableheadcolor}{gray}{0.95}  % 95% white → very light gray

\begin{table}[h!] 

\begin{center}
\caption{\label{tab:description}Number of teeth per class and overall distribution of tooth labels.}
\setlength{\tabcolsep}{10pt}
\renewcommand{\arraystretch}{1}

% ----------------------- GLOBAL STATS -----------------------
\begin{tabular}{lccc}
\toprule
 & Training & Testing & Total \\
\midrule
%\rowcolor{tableheadcolor} 
\# Patients & 600 & 300 & 900 \\
%\rowcolor{tableheadcolor} 
\# Scans    & 1200 & 600 & 1800 \\
%\rowcolor{tableheadcolor} 
\# Teeth    & 16004 & 7995 & 23999 \\
\bottomrule
\end{tabular}

\vspace{0.2cm}

% --------------------- UPPER / LOWER JAWS ---------------------
\begin{tabular}{lcc|lcc}
\toprule
\multicolumn{3}{c|}{\textbf{Upper Jaw}} &
\multicolumn{3}{c}{\textbf{Lower Jaw}} \\
\midrule
%\rowcolor{tableheadcolor}
Label & Training & Testing & Label & Training & Testing \\
\hline

18 & 28  & 14  & 38 & 31  & 16 \\
17 & 417 & 205 & 37 & 458 & 224 \\
16 & 592 & 298 & 36 & 586 & 296 \\
15 & 591 & 290 & 35 & 585 & 293 \\
14 & 586 & 293 & 34 & 589 & 292 \\
13 & 556 & 278 & 33 & 588 & 295 \\
12 & 591 & 297 & 32 & 600 & 300 \\
11 & 599 & 300 & 31 & 598 & 299 \\
21 & 599 & 300 & 41 & 598 & 300 \\
22 & 598 & 299 & 42 & 598 & 300 \\
23 & 561 & 283 & 43 & 589 & 297 \\
24 & 592 & 297 & 44 & 594 & 297 \\
25 & 584 & 286 & 45 & 581 & 289 \\
26 & 590 & 300 & 46 & 589 & 292 \\
27 & 425 & 213 & 47 & 456 & 225 \\
28 & 24  & 12  & 48 & 31  & 15 \\

%\rowcolor{tableheadcolor}
Total & 7933 & 3965 & Total & 8071 & 4030 \\

\bottomrule
\end{tabular}

\end{center}
\end{table}

\clearpage

\tableabvr~\ref{tab:description_landmark} summarizes the distribution of dental landmarks in the training and testing subsets. The distribution of dental landmarks across classes is highly consistent between the upper and lower jaws.

\begin{table}[hb!]
\begin{center}
\caption{\label{tab:description_landmark} Statistics on dental landmarks.}
\setlength{\tabcolsep}{10pt} % adjust horizontal spacing
\renewcommand{\arraystretch}{1}

% ----------------------- GLOBAL STATS -----------------------
\begin{tabular}{lccc}
\toprule
 & Training & Testing & Total \\
\midrule
%\rowcolor{tableheadcolor} 
\# Patients & 120 & 50 & 170 \\
%\rowcolor{tableheadcolor} 
\# Scans    & 240 & 100 & 340 \\
%\rowcolor{tableheadcolor} 
\# Landmarks & 21394 & 9142 & 30536 \\
\bottomrule
\end{tabular}

\vspace{0.2cm}

% --------------------- UPPER / LOWER JAWS ---------------------
\begin{tabular}{lcc|lcc}
\toprule
\multicolumn{3}{c|}{\textbf{Upper Jaw}} &
\multicolumn{3}{c}{\textbf{Lower Jaw}} \\
\midrule
%\rowcolor{tableheadcolor} 
Label & Training & Testing & Label & Training & Testing \\
\hline
Cusp    & 2583  & 1108 & Cusp    & 3010 & 1235 \\
Mesial  & 1581  & 680  & Mesial  & 1602 & 694  \\
Distal  & 1577  & 679  & Distal  & 1581 & 687  \\
Inner   & 1578  & 679  & Inner   & 1575 & 672  \\
Outer   & 1572  & 667  & Outer   & 1582 & 683  \\
Facial  & 1573  & 670  & Facial  & 1580 & 688  \\
%\rowcolor{tableheadcolor}
Total   & 10464 & 4483 & Total   & 10930 & 4659 \\
\bottomrule
\end{tabular}

\end{center}
\end{table}

\section{Experimental results}

Teeth3DS+ is introduced to support the MICCAI Challenges focusing on 3D dental scans analysis, along with a standardized evaluation protocol to assess competing approaches. In the following, we present the obtained experimental results for both tasks.

\subsection{Teeth segmentation and labeling}

The 3DTeethSeg challenge attracted numerous teams, but only six were retained for the final phase: CGIP, FiboSeg, IGIP, TeethSeg, OS, and Radboud. Technical details of their methods are provided in \cite{ben20233dteethseg}.

\paragraph{\textbf{Metrics}} Evaluating teeth segmentation and labeling requires specialized metrics. We proposed three dedicated metrics, which have since been adopted by the community for consistent benchmarking:
\begin{itemize}
    \item Teeth Localization Accuracy (TLA): This metric calculates the mean normalized Euclidean distance between each ground-truth tooth centroid and the nearest predicted centroid. Each distance is normalized by the size of the corresponding ground-truth tooth. When a centroid is missing, for example, due to a method failure or an incomplete output, a fixed penalty of 5 is assigned for that ground-truth tooth, equivalent to five times its actual size. Since patients may have different numbers of teeth, the final score is obtained by averaging over all ground-truth teeth across both test sets.
    
    \item  Teeth Identification Rate (TIR): This metric represents the percentage of correctly identified teeth relative to all ground-truth teeth in the two test sets. A ground-truth tooth is counted as correctly identified when the nearest predicted centroid lies within half the size of the corresponding ground-truth tooth and is assigned the same label.

    \item Teeth Segmentation Accuracy (TSA): This metric is computed as the average F1-score over all tooth point cloud instances, with each tooth's F1-score calculated from its precision and recall. 

\end{itemize}  

\begin{table}[b!]
\renewcommand{\arraystretch}{1.3}
\setlength{\tabcolsep}{25pt} % increase horizontal spacing

\centering
\begin{tabular}{@{}lccc@{}}
\toprule
\multirow{2}{*}{\textbf{Team}} 
& \multicolumn{3}{c}{\textbf{Metrics}} \\
\cmidrule(lr){2-4}
& \textbf{TLA} & \textbf{TSA} & \textbf{TIR} \\
\midrule

CGIP      & 0.9658 & 0.9859 & 0.9100 \\
FiboSeg   & 0.9924 & 0.9293 & 0.9223 \\
IGIP      & 0.9244 & 0.9750 & 0.9289 \\
TeethSeg  & 0.9184 & 0.9678 & 0.8538 \\
OS        & 0.7845 & 0.9693 & 0.8940 \\
Radboud   & 0.6242 & 0.8886 & 0.8795 \\
\bottomrule
\end{tabular}
\caption{Evaluation metrics for the participating teams in the MICCAI 3DTeethSeg challenge.}
\label{tab:results_3DTeethSeg}
\end{table}

\paragraph{\textbf{Obtained results}} \tableabvr~\ref{tab:results_3DTeethSeg} summarizes the achieved evaluation metrics for the participating teams in 3DTeethSeg. The methods obtained a teeth localization accuracy between $62.42\%$  and $99.24\%$, a teeth segmentation accuracy between $88.86\%$ and $98.59\%$, and a teeth identification rate between $85.38\%$ and $92.89\%$.

\begin{figure*}[t!]
    \centering

    % -------- Row 1 (single GT figure) --------
    \begin{minipage}{0.32\textwidth}
        \centering
        \includegraphics[width=\textwidth]{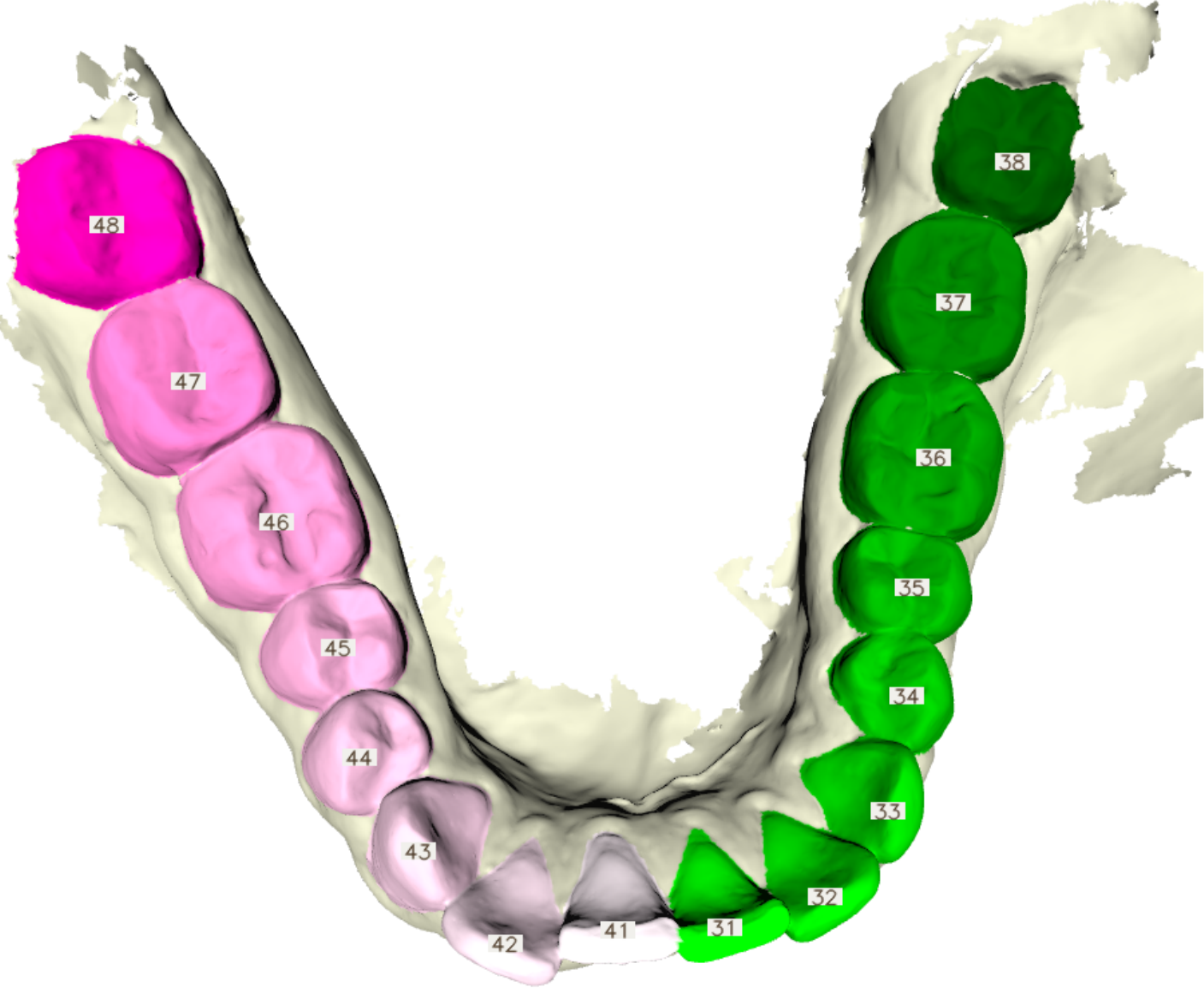}
        \caption*{(a) Ground-truth}
    \end{minipage}

    \vspace{0.4cm}

    % -------- Row 2 (3 figures) --------
    \begin{minipage}{0.32\textwidth}
        \centering
        \includegraphics[width=\textwidth]{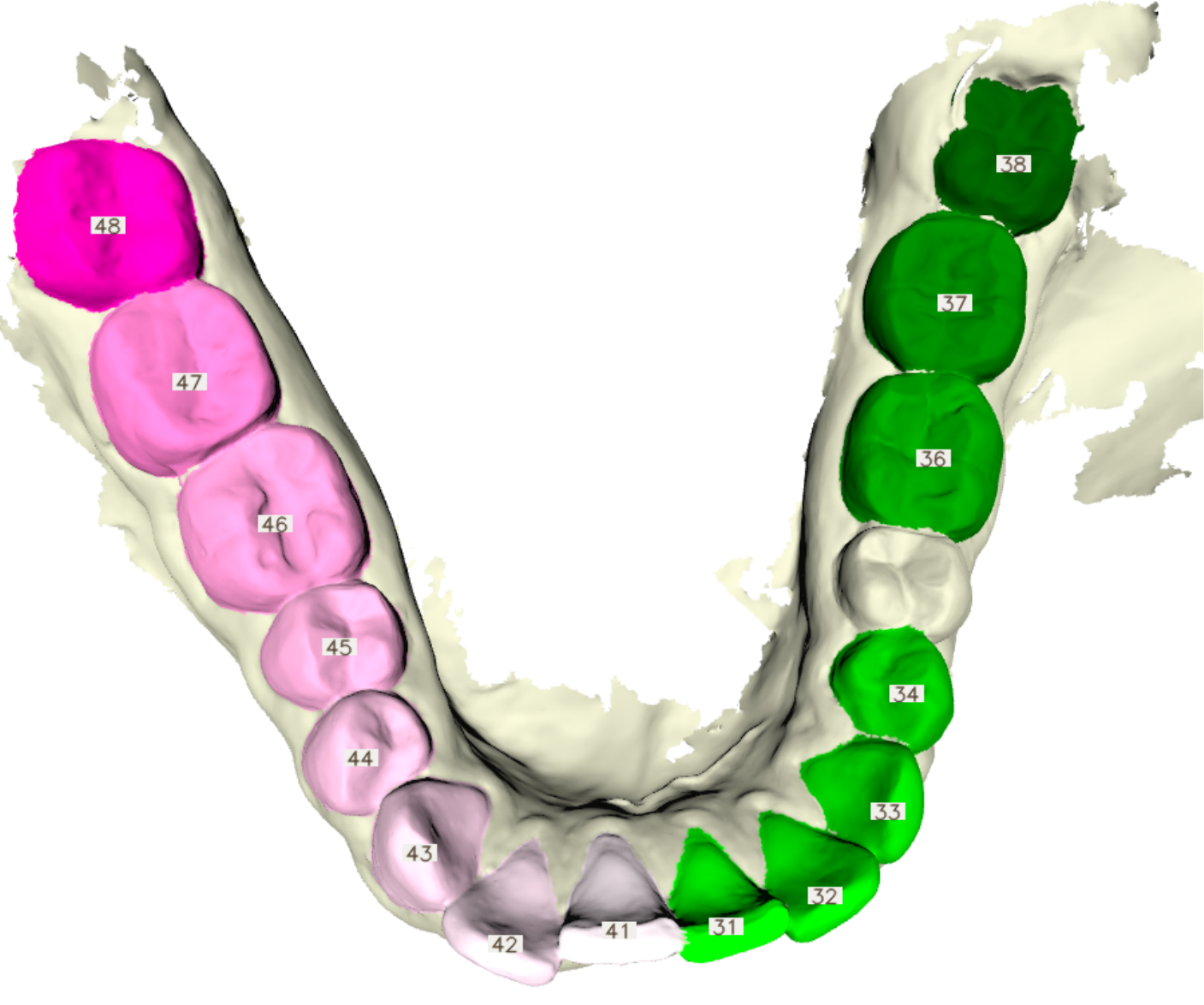}
        \caption*{(b) OS team}
    \end{minipage}
    \begin{minipage}{0.32\textwidth}
        \centering
        \includegraphics[width=\textwidth]{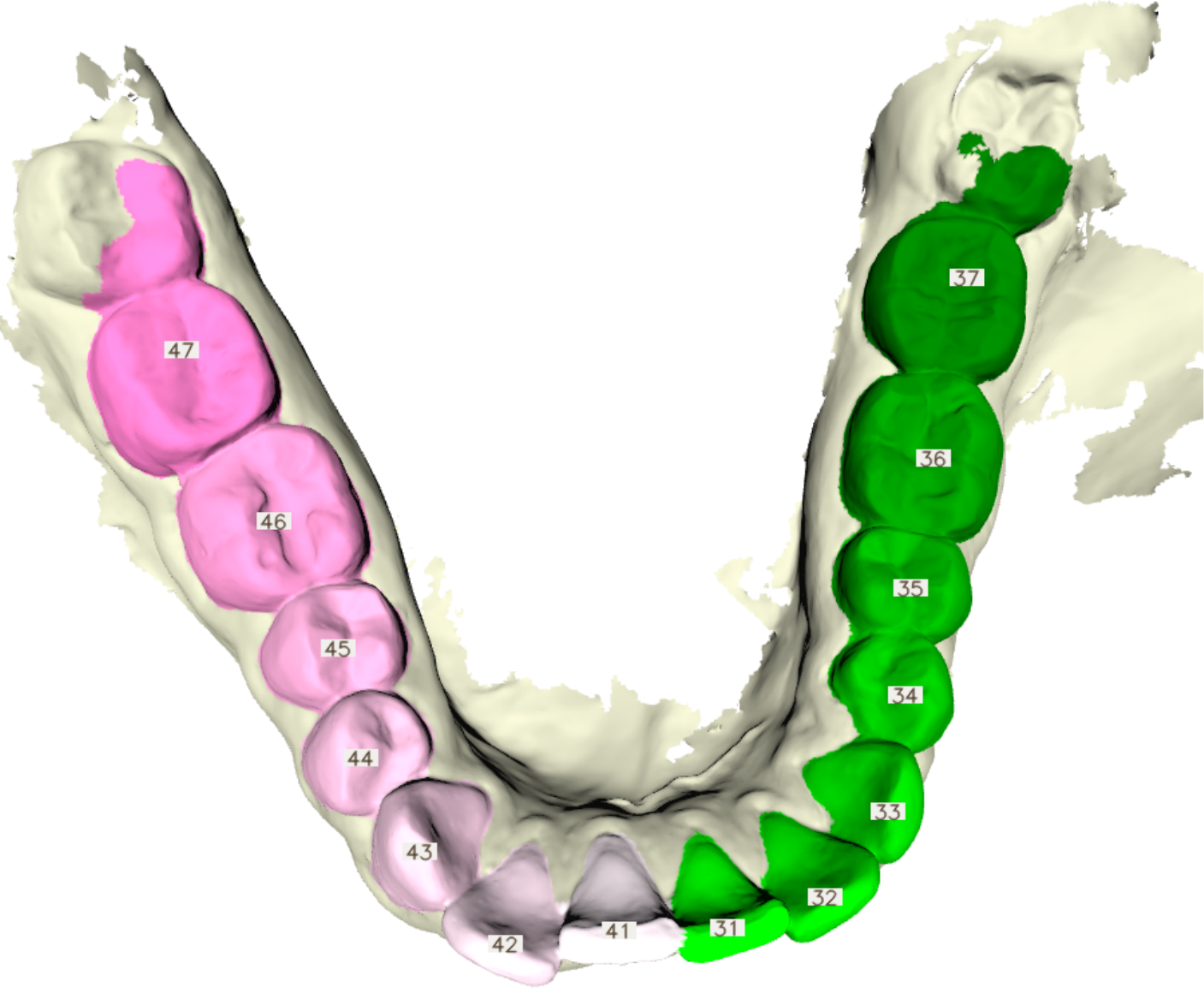}
        \caption*{(c) TeethSeg team}
    \end{minipage}
    \begin{minipage}{0.32\textwidth}
        \centering
        \includegraphics[width=\textwidth]{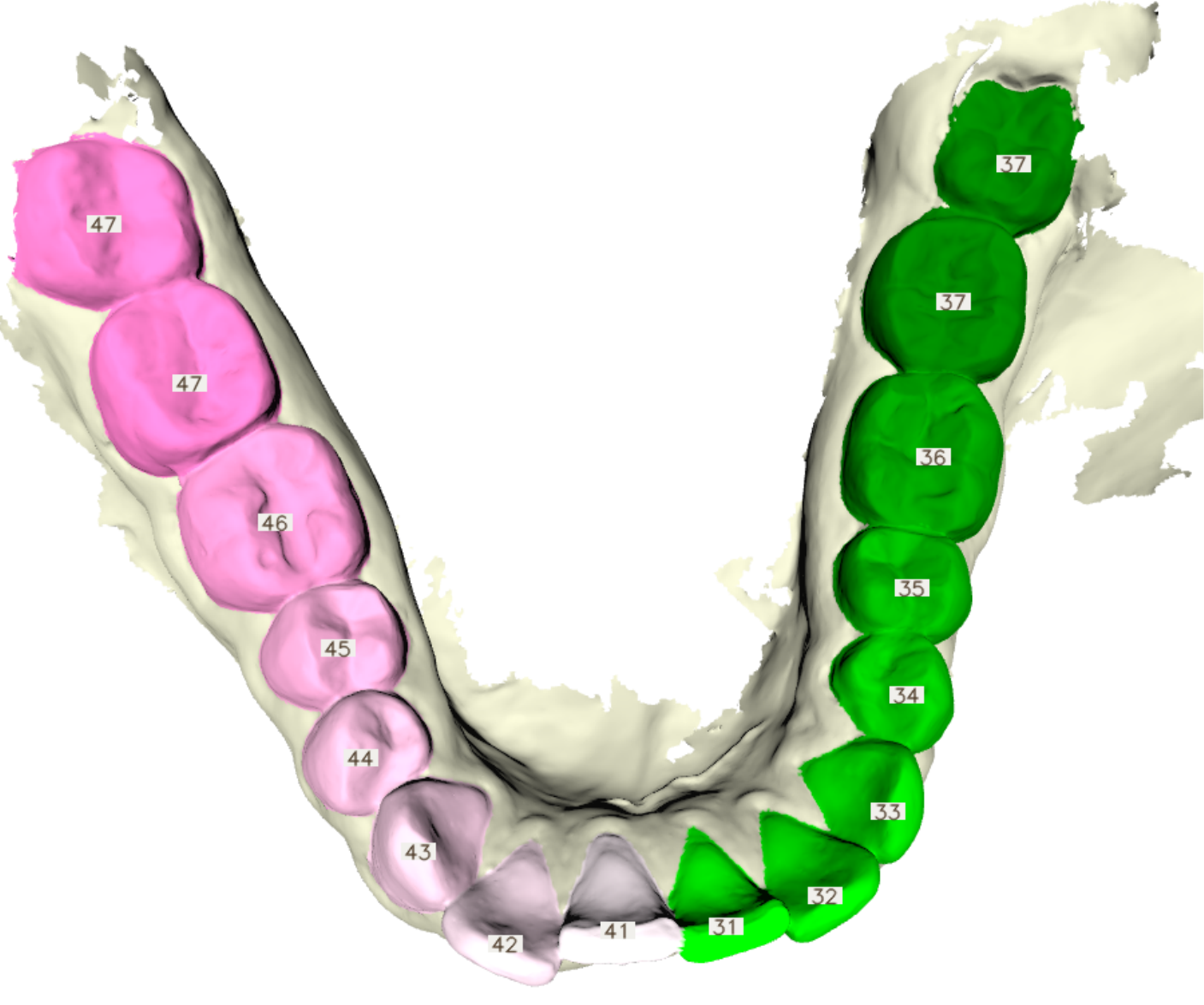}
        \caption*{(d) CGIP team}
    \end{minipage}

    \vspace{0.4cm}

    % -------- Row 3 (3 figures) --------
    \begin{minipage}{0.32\textwidth}
        \centering
        \includegraphics[width=\textwidth]{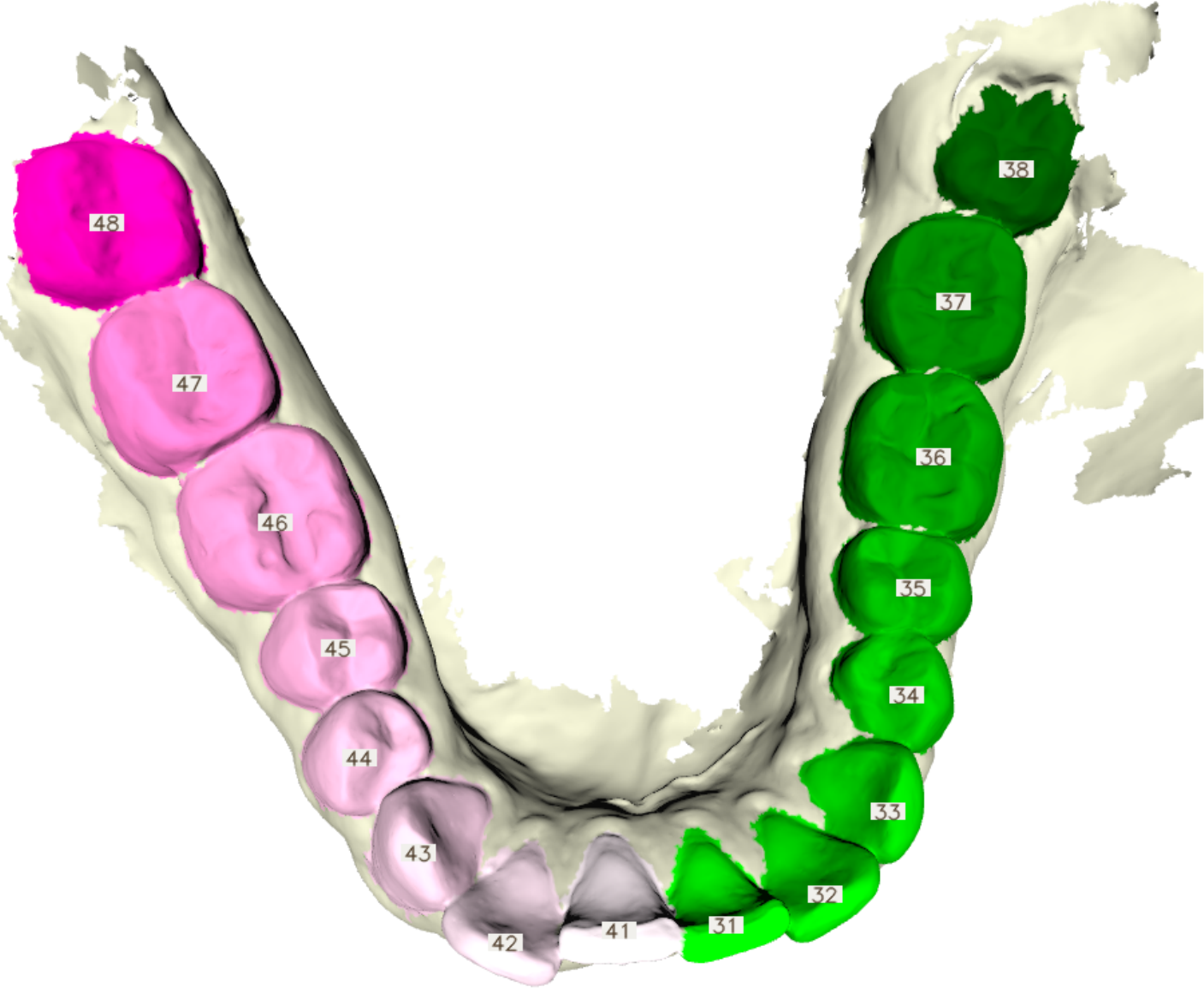}
        \caption*{(e) IGIP team}
    \end{minipage}
    \begin{minipage}{0.32\textwidth}
        \centering
        \includegraphics[width=\textwidth]{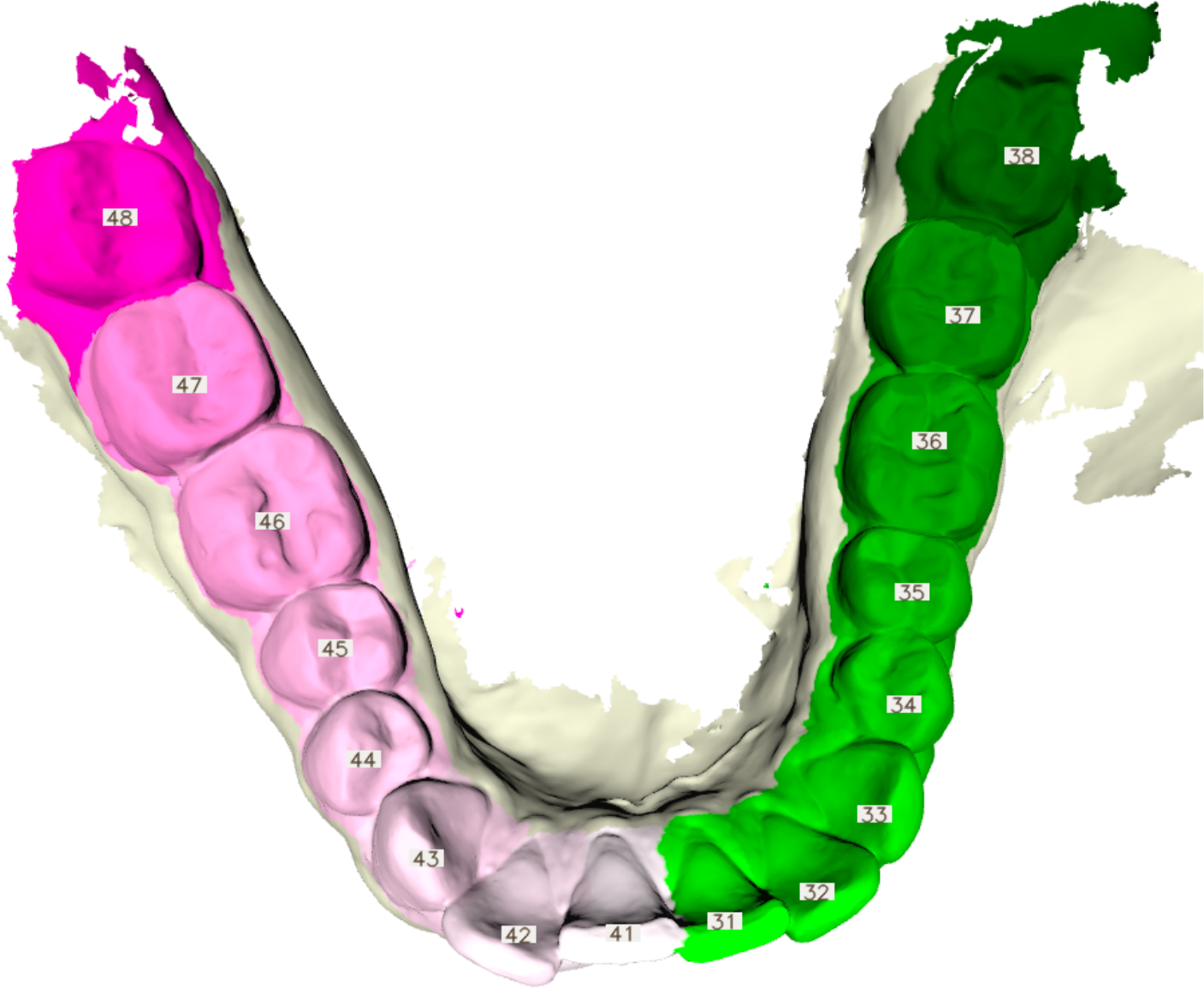}
        \caption*{(f) FiboSeg team}
    \end{minipage}
    \begin{minipage}{0.32\textwidth}
        \centering
        \includegraphics[width=\textwidth]{OS_87N5YSES_lower_pred_cropped.pdf}
        \caption*{(g) Radboud team}
    \end{minipage}

    \caption{Example of visual comparison of the obtained results in the segmentation task.}
    \label{fig:visual_comp_challenge1}
\end{figure*}

\figureabvr \ref{fig:visual_comp_challenge1} shows a visual representation of the segmentation results obtained by the competing methods. Some approaches produce clean and anatomically coherent boundaries, while others struggle particularly in challenging regions. This variability underscores the difficulty of the dataset, reinforces its value as a robust benchmark, and motivates future research toward more accurate and reliable segmentation and labeling methods.

\begin{figure}[t]
    \centering
    % First row: two subfigures
    \begin{subfigure}[b]{0.49\textwidth}
        \centering
        \includegraphics[width=\textwidth]{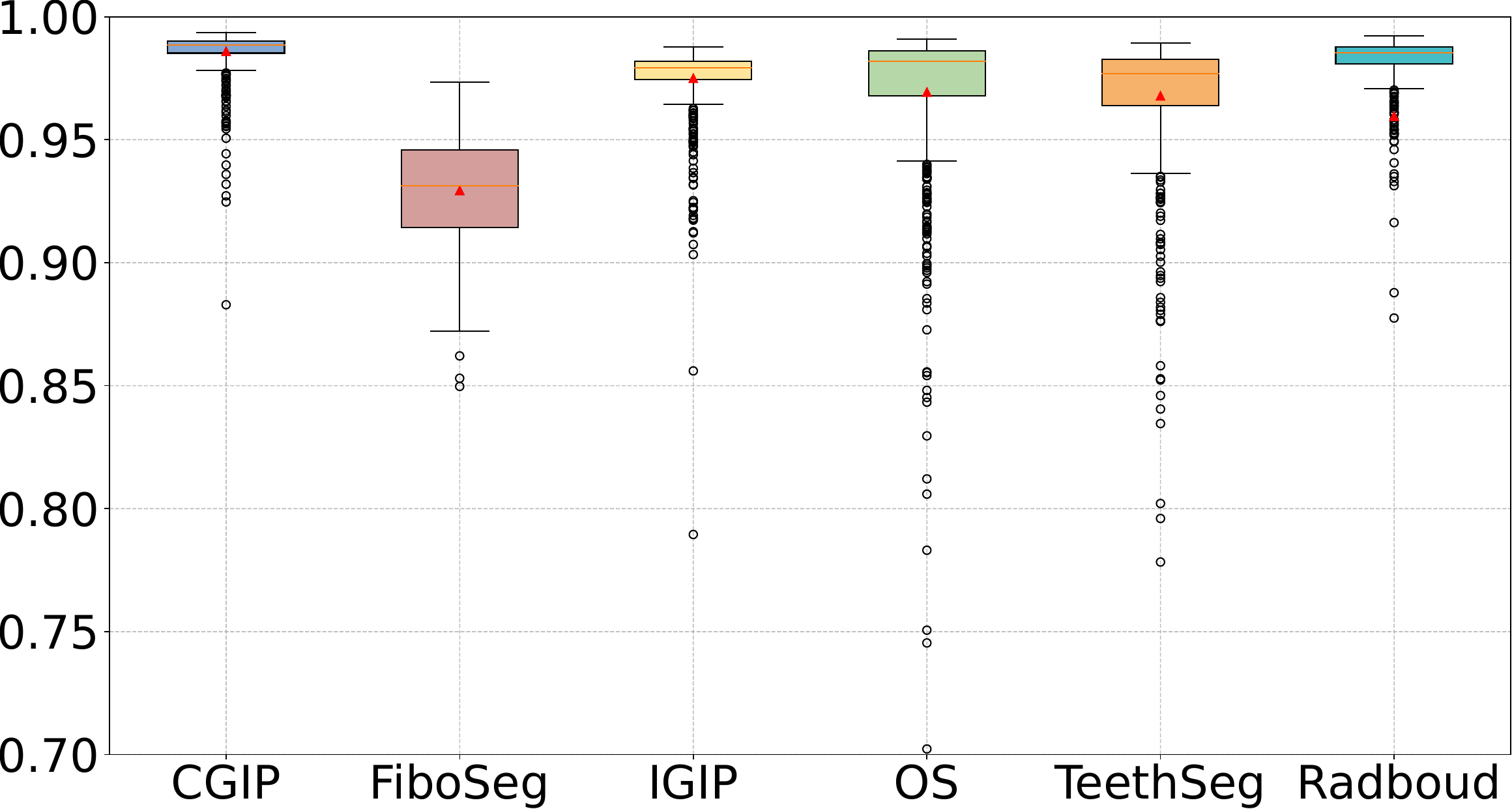}
        \caption{TSA}
        \label{fig:fig1}
    \end{subfigure}
    \hfill
    \begin{subfigure}[b]{0.49\textwidth}
        \centering
        \includegraphics[width=\textwidth]{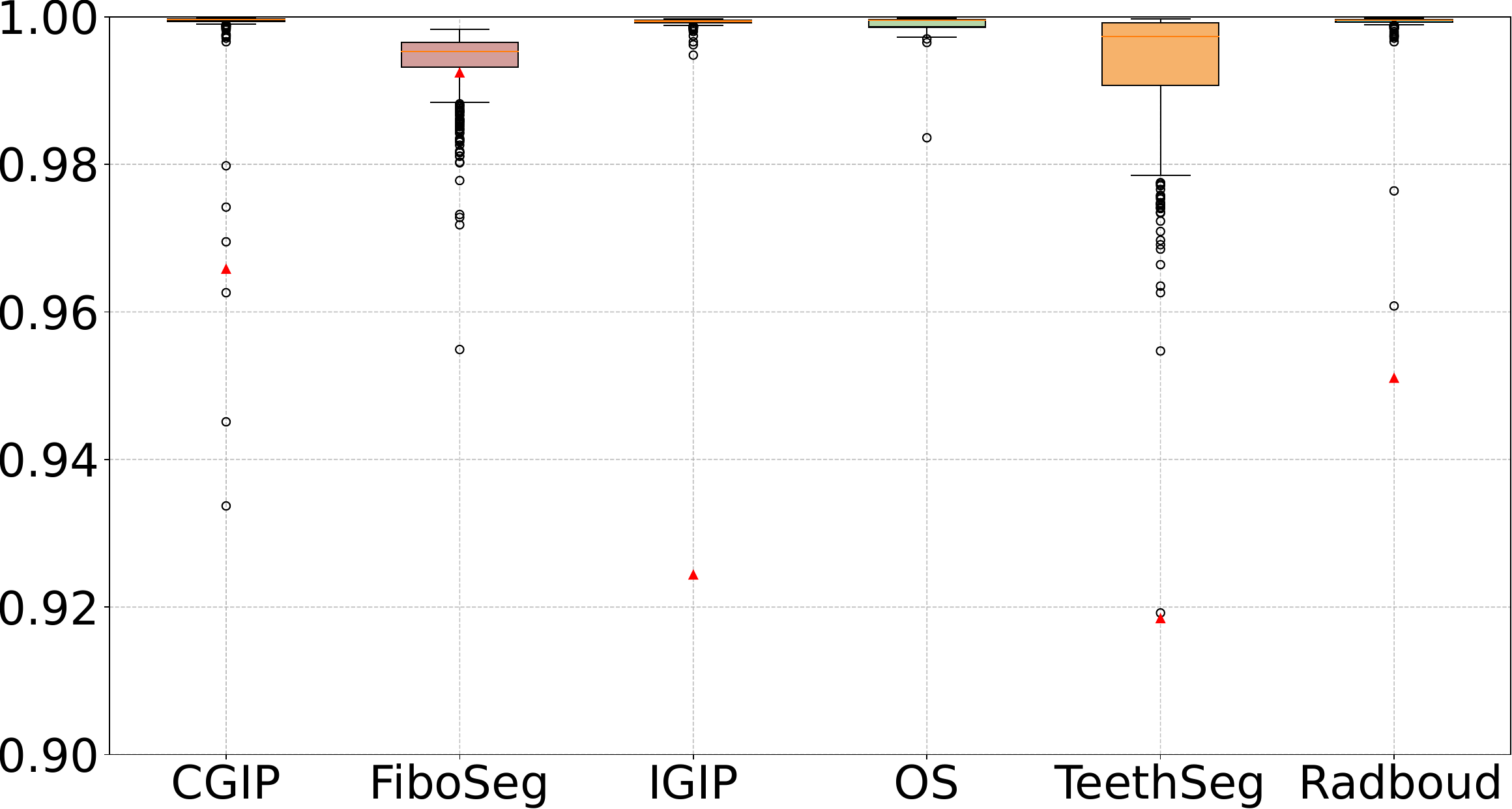}
        \caption{TLA}
        \label{fig:fig2}
    \end{subfigure}

    % Second row: one centered subfigure
    \vspace{0.5cm} % space between rows
    \begin{subfigure}[b]{0.49\textwidth}
        \centering
        \includegraphics[width=\textwidth]{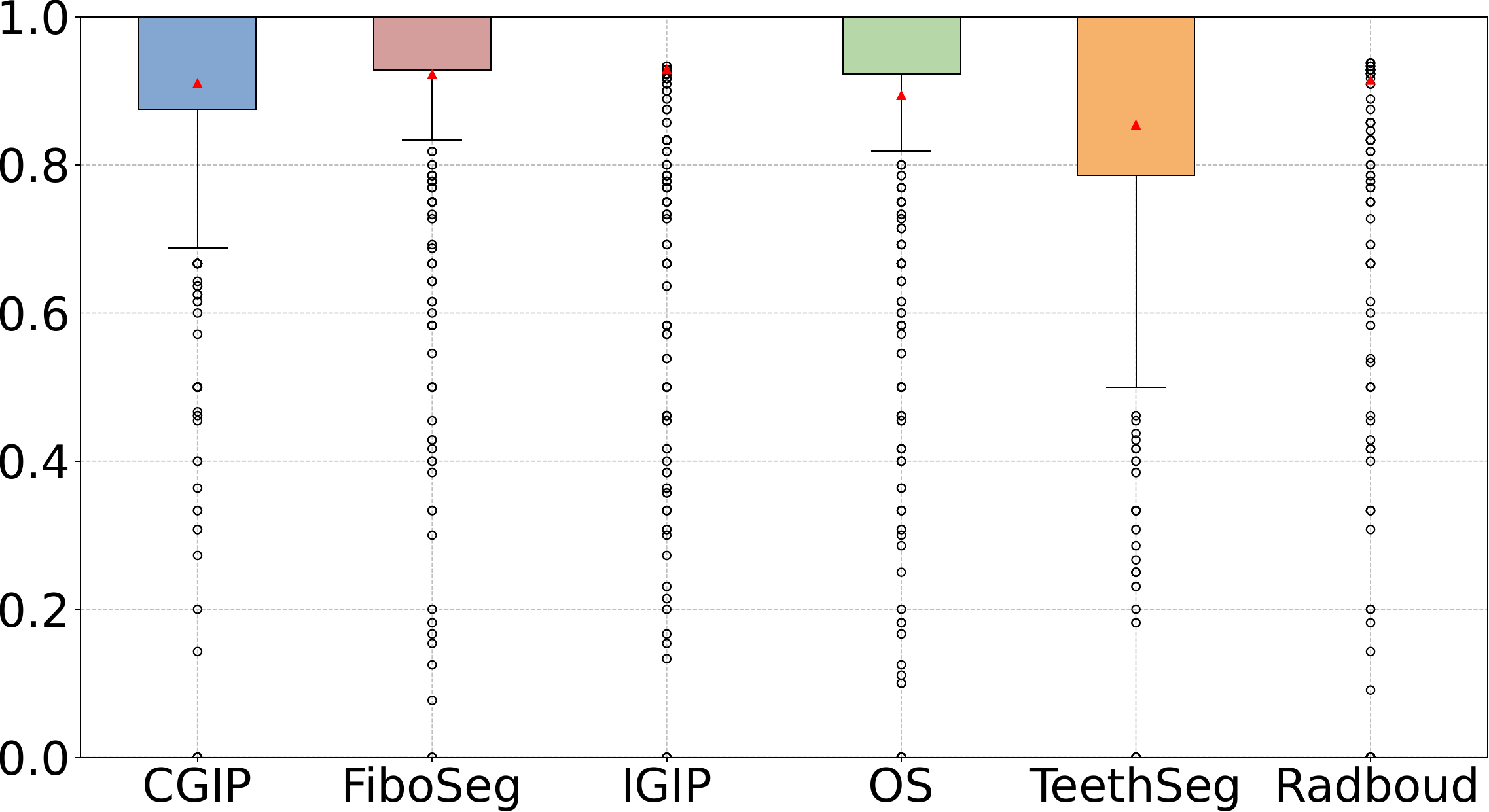}
        \caption{TIR}
        \label{fig:fig3}
    \end{subfigure}

    \caption{Boxplot distribution of TSA, TLA, and TIR across segmentation methods.}
    \label{fig:boxplot_challenge1}
\end{figure}

\figureabvr \ref{fig:boxplot_challenge1} presents boxplots of TLA, TIR, and TSA for each scan in the segmentation task. Some methods show wider boxes, indicating higher variability across scans, while others are more consistent. The outliers generally correspond to challenging clinical cases, such as irregular anatomy or artifacts. Although they reduce the overall statistics, they reflect the heterogeneity of data. They are essential for evaluating the robustness of segmentation methods and ensuring that the dataset captures realistic clinical variability.

%\clearpage
\subsection{Teeth landmark identification}
The 3DTeethLand challenge initially attracted several participating teams. However, only six were selected for the final phase: Radboud, ChohoTech, YY-LAB, YN-LAB, IGIP-LAB, and 3DIMLAND. The technical details of the proposed methods are provided in \cite{benhamadou2025detectingdentallandmarksintraoral}. 

\paragraph{\textbf{Metrics}}

Standard metrics such as root mean square error (RMSE) and mean absolute error (MAE) are commonly used to assess landmark localization accuracy, typically under the assumption of a fixed number of landmarks. However, the task involves detecting a variable number of landmarks per tooth. Therefore, we consider two additional metrics that are better suited to this setting and more aligned with the characteristics of landmark identification. Landmarks are grouped into four clinically relevant categories: mesial/distal, cusps, outer/inner, and facial.

\begin{itemize}
    \item Mean Average Precision (mAP): The metric is computed by averaging the Average Precision (AP) across each category, with AP derived from the area under the precision–recall curve at multiple distance thresholds.
    \item Mean Average Recall (mAR): This metric aggregates the Average Recall (AR) per category, calculated from the area under the recall–exp(distance) curve.
\end{itemize}
A prediction is considered correct if the Euclidean distance to the reference landmark is below a predefined threshold, and multiple thresholds are used to evaluate performance across different clinical requirements.

The localization criterion relies on the Euclidean distance between each predicted landmark and its corresponding ground-truth point. A prediction is marked as a correct detection when this distance is below a chosen threshold. Since clinical tolerance levels differ, performance is evaluated across a series of thresholds spanning from 0 mm to 3 mm in steps of 0.1 mm, providing a comprehensive view of the method's behavior under various practical conditions.

\paragraph{\textbf{Obtained results}}
\tableabvr~\ref{tab:challenge2_result} presents the evaluation metrics obtained by the participating teams in 3DTeethLand. Across all methods, mAP values range from 0.578 to 0.785, while mAR values range from 0.438 to 0.656. The relative consistency of these results across teams indicates that the dataset provides rich and well-structured geometric information, supporting accurate and stable model performance and confirming its suitability as a benchmark for dental landmark identification. 

\begin{figure*}[h!]
    \centering

    \begin{minipage}{0.32\textwidth}
        \centering
        \includegraphics[width=\textwidth]{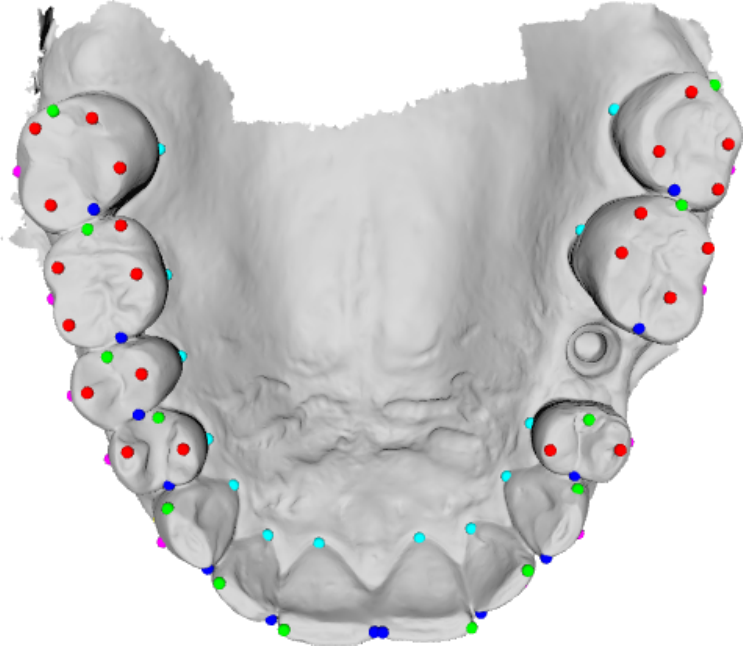}
        \caption*{(a) Ground-truth}
    \end{minipage}

    \vspace{0.38cm}

    \begin{minipage}{0.32\textwidth}
        \centering
        \includegraphics[width=\textwidth]{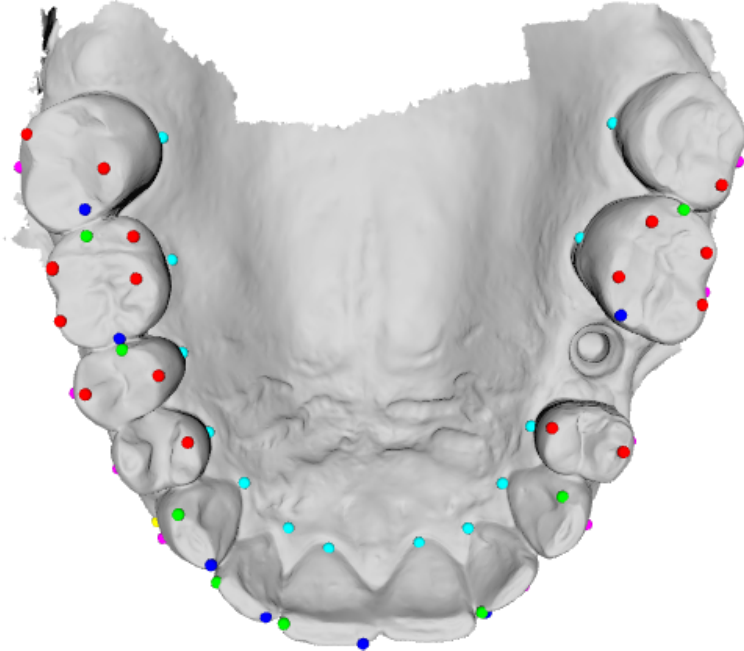}
        \caption*{(b) 3DIMLAND team}
    \end{minipage}
    \begin{minipage}{0.32\textwidth}
        \centering
        \includegraphics[width=\textwidth]{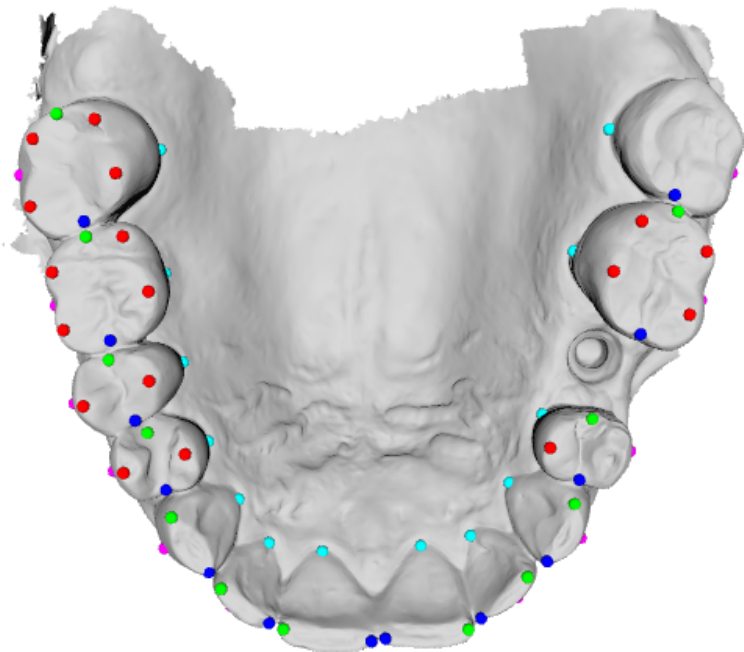}
        \caption*{(c) Chohotech team}
    \end{minipage}
    \begin{minipage}{0.32\textwidth}
        \centering
        \includegraphics[width=\textwidth]{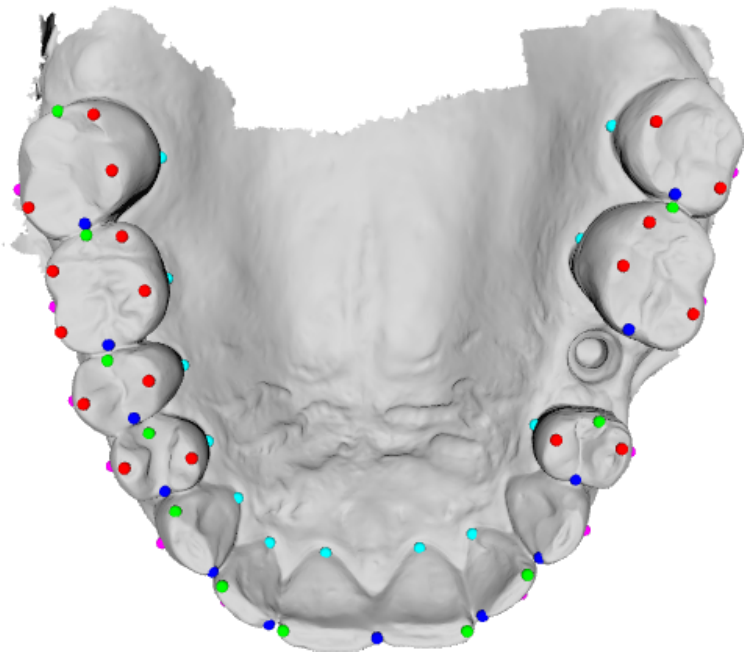}
        \caption*{(d) YN-LAB team}
    \end{minipage}

    \vspace{0.38cm}

    \begin{minipage}{0.32\textwidth}
        \centering
        \includegraphics[width=\textwidth]{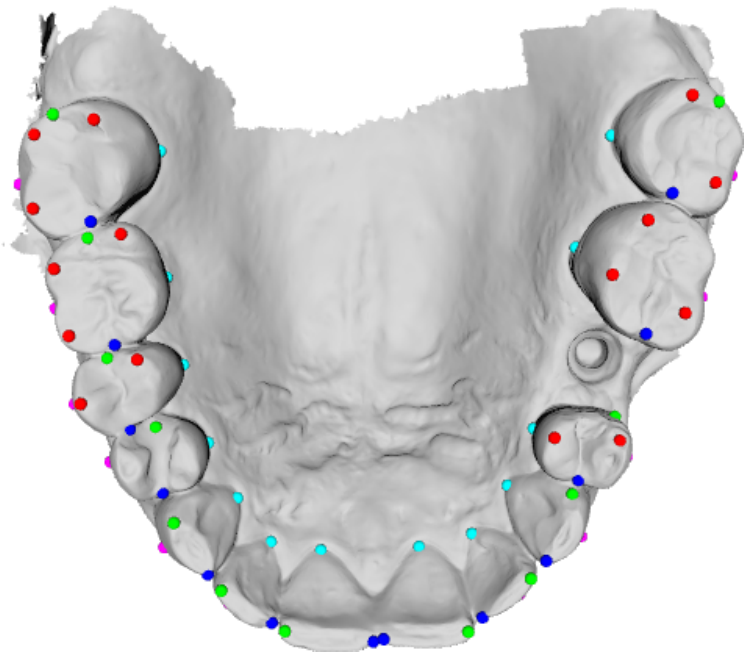}
        \caption*{(e) YY-LAB team}
    \end{minipage}
    \begin{minipage}{0.32\textwidth}
        \centering
        \includegraphics[width=\textwidth]{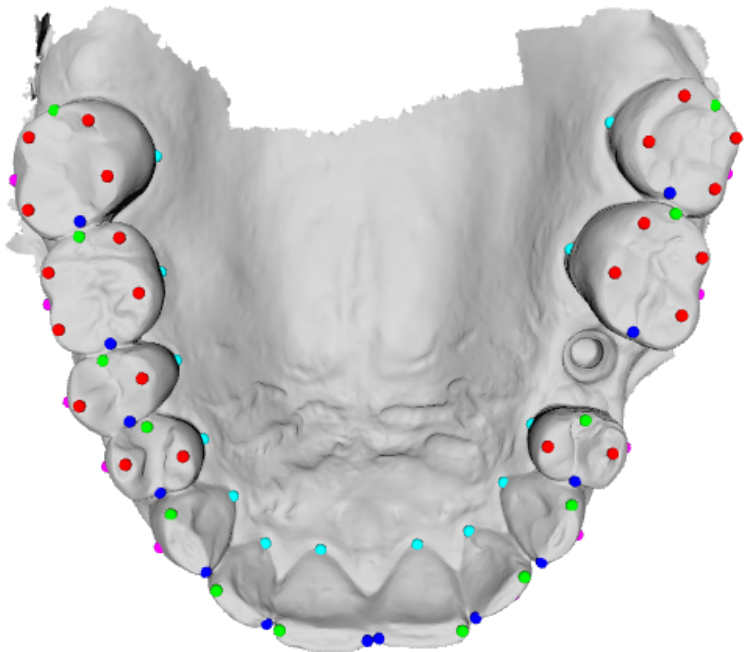}
        \caption*{(f) Radboud team}
    \end{minipage}
    \begin{minipage}{0.32\textwidth}
        \centering
        \includegraphics[width=\textwidth]{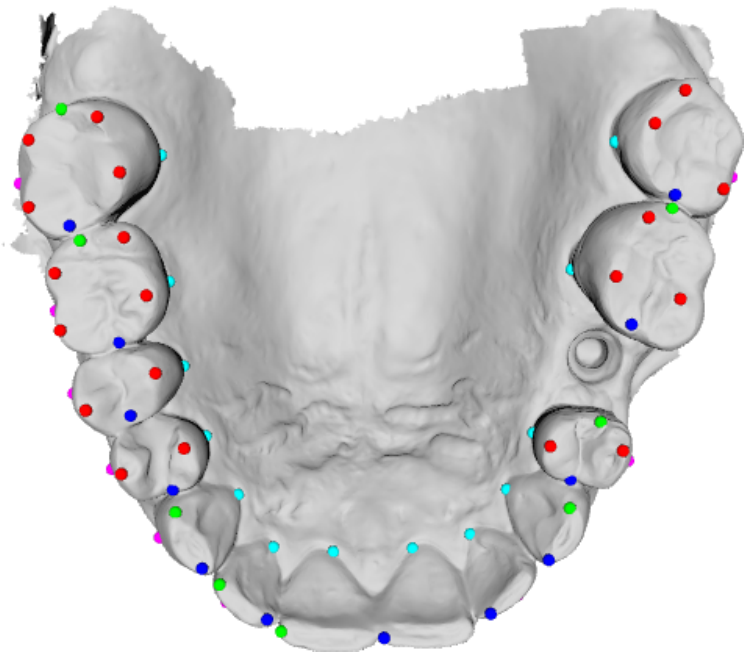}
        \caption*{(g) IGIP-LAB team}
    \end{minipage}

    \caption{Example of visual comparison of the landmark identification task.}
    \label{fig:visual_comparison_challenge2}
\end{figure*}

\begin{figure*}[t]
    \centering
    \begin{adjustbox}{max width=0.9\textwidth} % scales entire figure nicely
        \begin{minipage}{\textwidth}
            \begin{subfigure}[t]{0.49\textwidth}
                \centering
                \includegraphics[width=\textwidth]{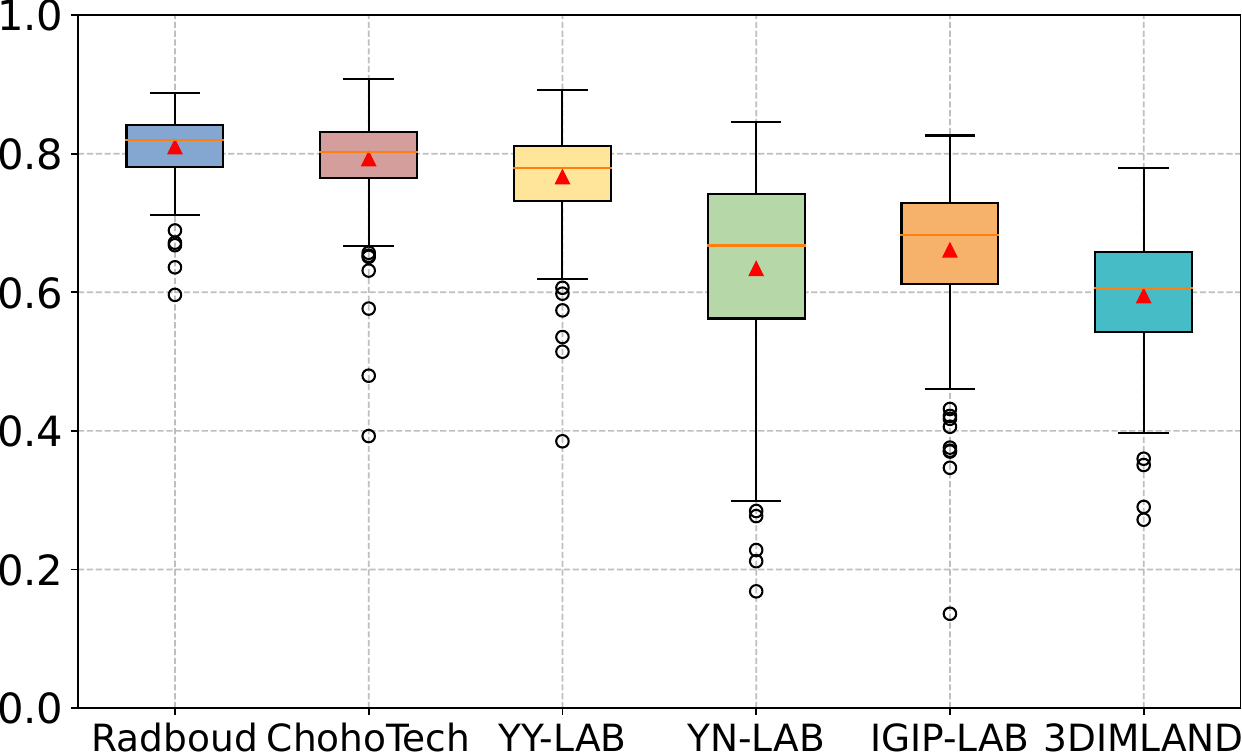}
                \caption{mAP for mesial\_distal category}
                \label{fig:mesial_boxplot_mAP}
            \end{subfigure}\hfill
            \begin{subfigure}[t]{0.49\textwidth}
                \centering
                \includegraphics[width=\textwidth]{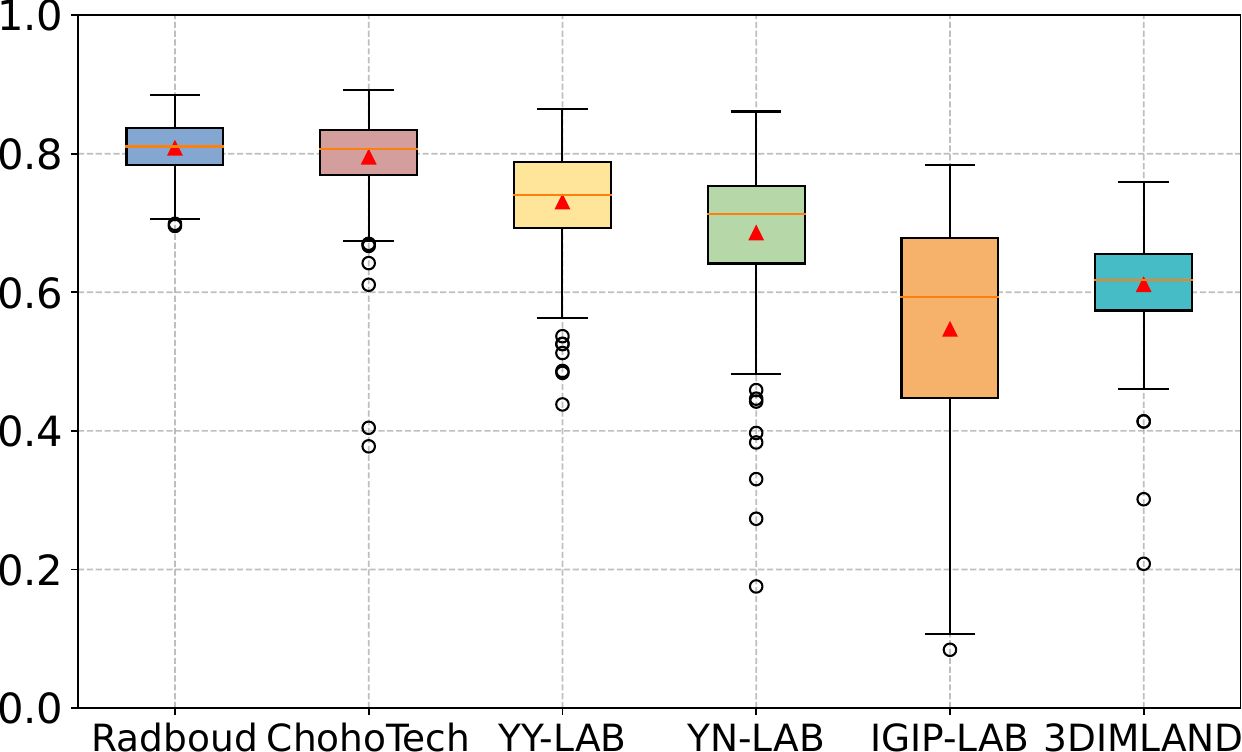}
                \caption{mAP for inner\_outer category}
                \label{fig:distal_boxplot_mAP}
            \end{subfigure}

            \vspace{0.4cm}

            \begin{subfigure}[t]{0.49\textwidth}
                \centering
                \includegraphics[width=\textwidth]{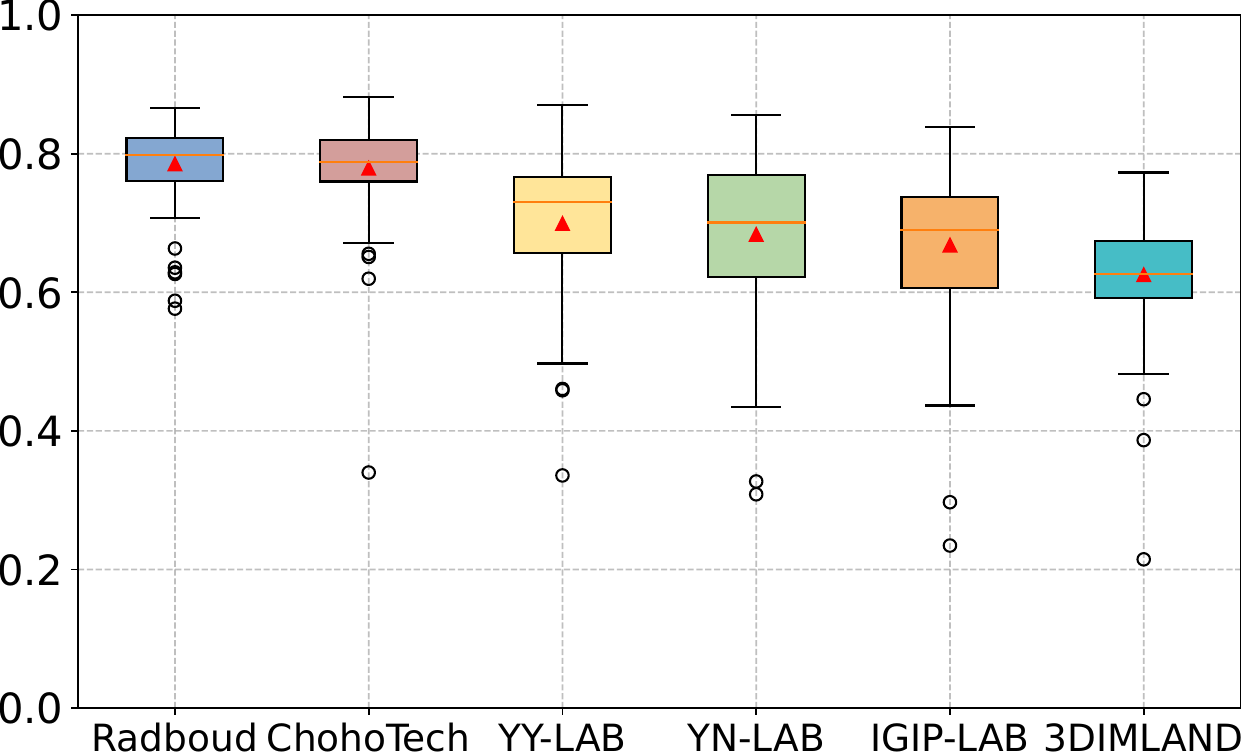}
                \caption{mAP for cusp category}
                \label{fig:cusp_boxplot_mAP}
            \end{subfigure}\hfill
            \begin{subfigure}[t]{0.49\textwidth}
                \centering
                \includegraphics[width=\textwidth]{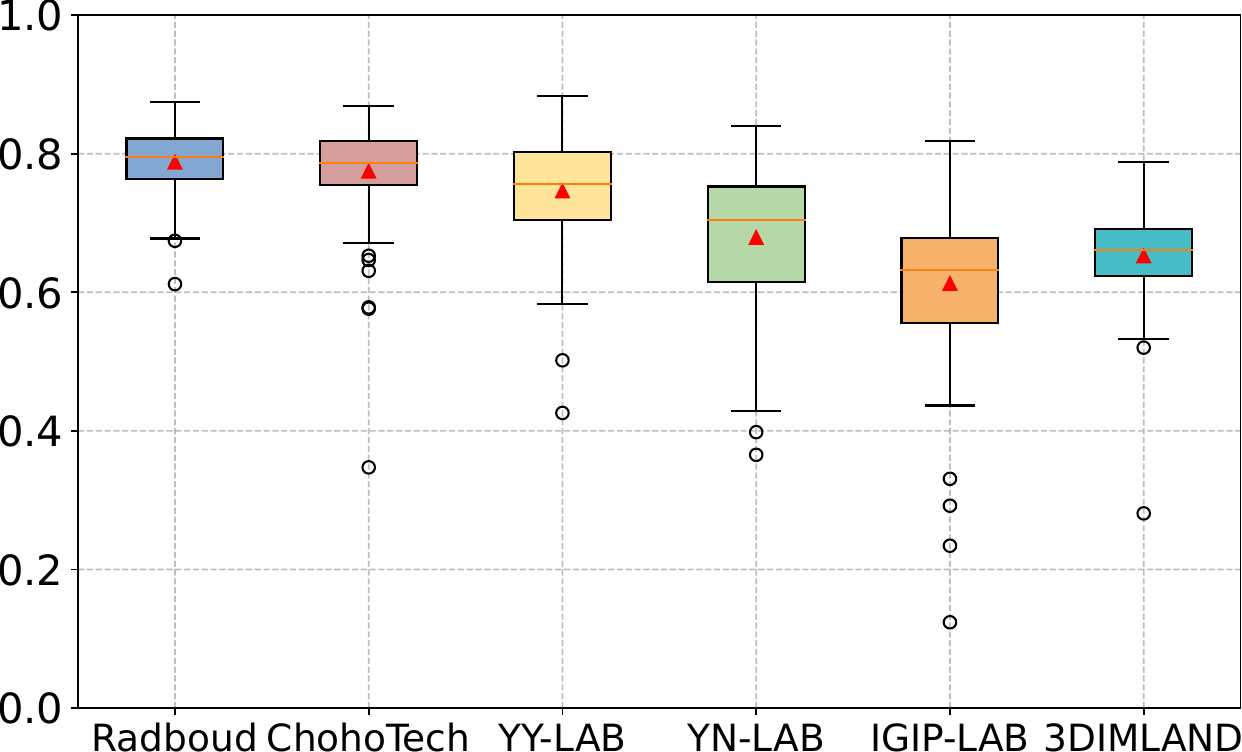}
                \caption{mAP for facial category}
                \label{fig:facialpoint_boxplot_mAP}
            \end{subfigure}

        \end{minipage}
    \end{adjustbox}
    \caption{mAP score for each landmark category per scan.}
    \label{fig:mAP_boxplot}
\end{figure*}

\begin{figure*}[h!]
    \centering
    \resizebox{0.9\textwidth}{!}{%
        \begin{minipage}{\textwidth}
            % First row
           
            \begin{subfigure}[t]{0.49\textwidth}
                \centering
                \includegraphics[width=\textwidth]{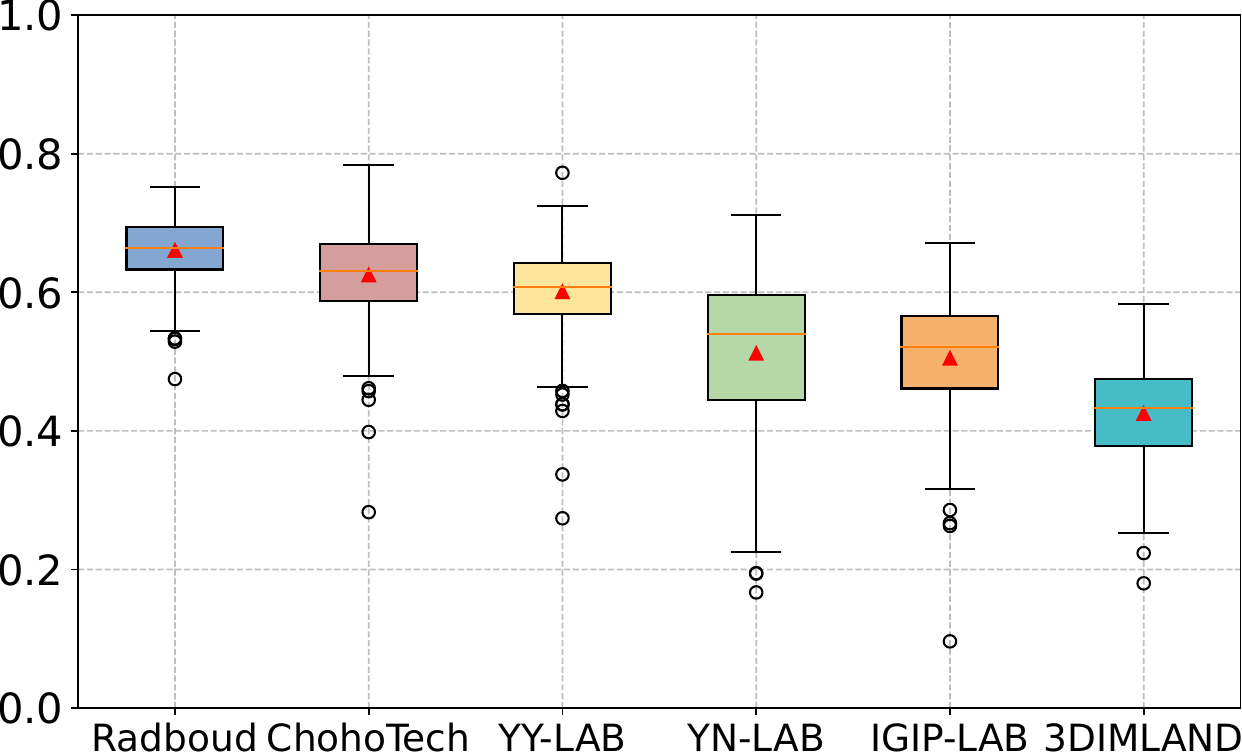}
                \caption{mAR for mesial\_distal category}
                \label{fig:mesial_distal_boxplot_mAR}
            \end{subfigure}\hfill
            \begin{subfigure}[t]{0.49\textwidth}
                \centering
                \includegraphics[width=\textwidth]{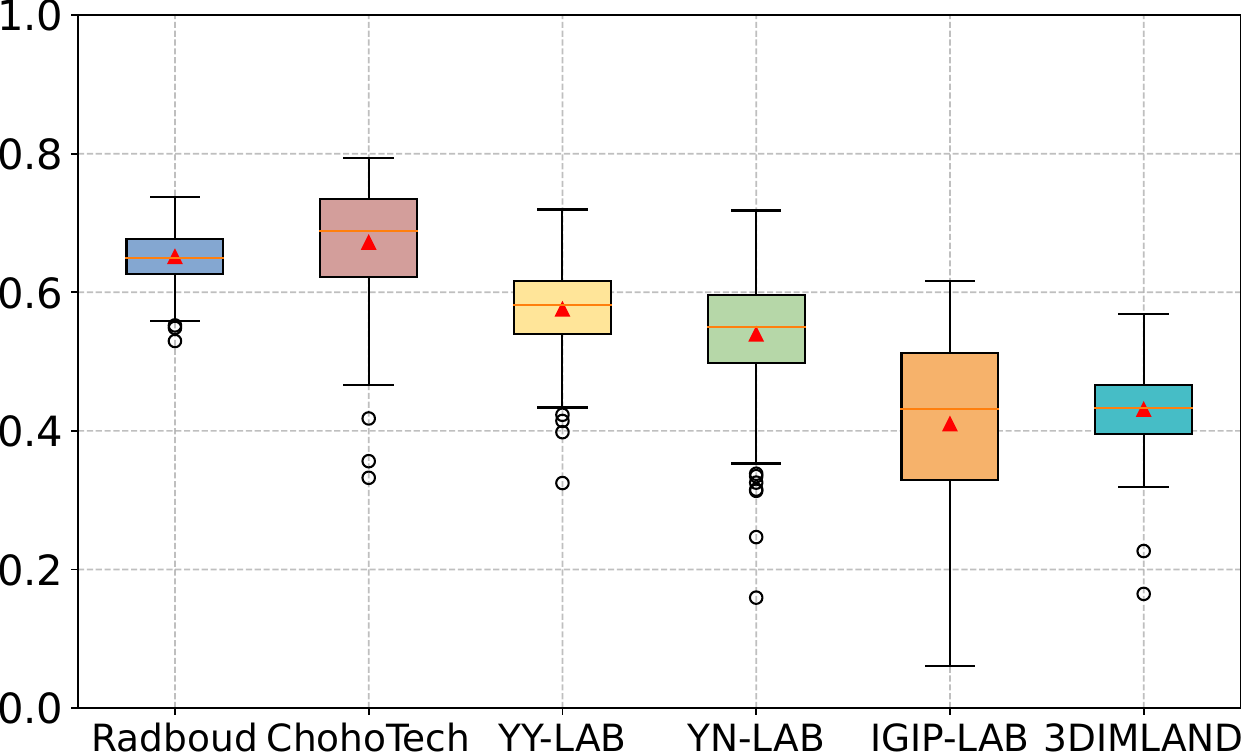}
                \caption{mAR for inner\_outer category}
                \label{fig:inner_outer_boxplot_mAR}
            \end{subfigure}
            \vspace{0.4cm} % space between rows

            % Second row
 \begin{subfigure}[t]{0.49\textwidth}
                \centering
                \includegraphics[width=\textwidth]{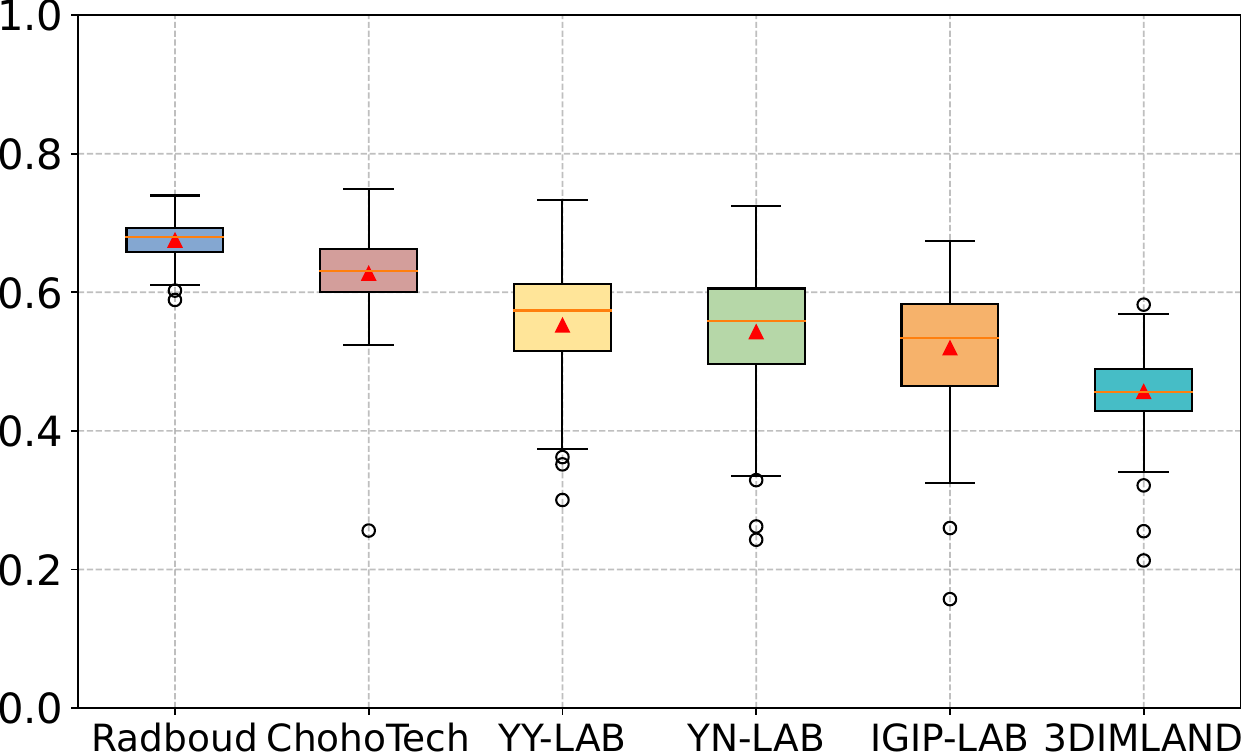}
                \caption{mAR for cusp category}
                \label{fig:cusp_boxplot_mAR}
            \end{subfigure}\hfill
            \begin{subfigure}[t]{0.49\textwidth}
                \centering
                \includegraphics[width=\textwidth]{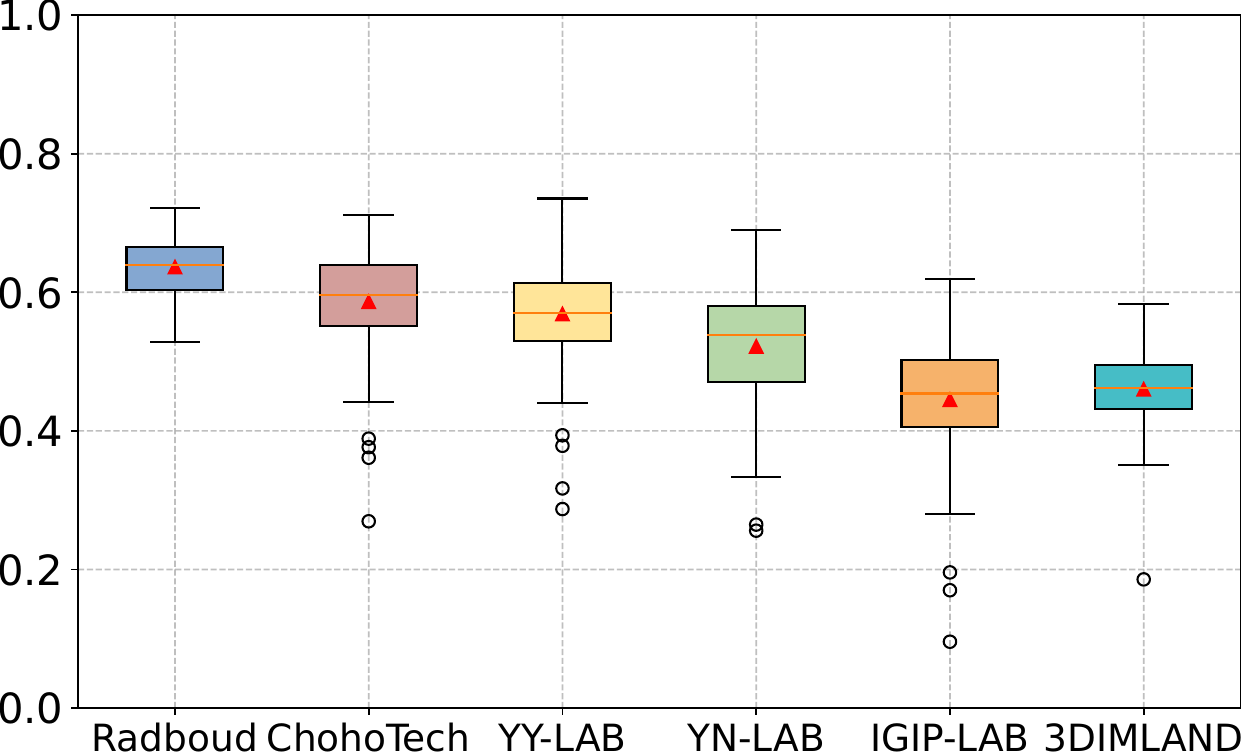}
                \caption{mAR for facial category}
                \label{fig:facialpoint_boxplot_mAR}
            \end{subfigure}

        \end{minipage}%
    }
    \caption{mAR score for each landmark category per scan.}
    \label{fig:mAR_boxplot}
\end{figure*}

The visual comparison of landmark identification in \figureabvr \ref{fig:visual_comparison_challenge2} across the six methods shows that some teams, such as Radboud, achieve precise and consistent localization, while others, such as 3DIMLAND, exhibit missing or inaccurately placed landmarks. Figures \ref{fig:mAP_boxplot} and \ref{fig:mAR_boxplot} illustrate the distribution of mAP and mAR values across scans and categories. Overall, these results demonstrate the effectiveness of our dataset in enabling robust model evaluation. While some teams achieve consistently high performance, the observed outliers represent challenging cases, reflecting the difficulty of the dataset and highlighting opportunities for future methods to further improve accuracy and robustness.

\begin{table*}[t!]
\renewcommand{\arraystretch}{1.3}
\centering
\resizebox{\textwidth}{!}{
\begin{tabular}{@{}lcccccccccc@{}}
\toprule
\multirow{2}{*}{\textbf{Team}} 
& \multicolumn{4}{c}{\textbf{mAP}} 
& \multicolumn{4}{c}{\textbf{mAR}} 
& \multirow{2}{*}{\begin{tabular}[c]{@{}c@{}}\textbf{Overall} \\  \textbf{mAP} \end{tabular}}
& \multirow{2}{*}{   \begin{tabular}[c]{@{}c@{}}\textbf{Overall} \\  \textbf{mAP} \end{tabular}  } \\
\cmidrule(lr){2-5} \cmidrule(lr){6-9}
& \textbf{mesial\_distal} & \textbf{inner\_outer} & \textbf{cusp} & \textbf{facial}
& \textbf{mesial\_distal} & \textbf{inner\_outer} & \textbf{cusp} & \textbf{facial} \\
\midrule

Radboud 
& 0.792 & 0.793 & 0.772 & 0.768
& 0.651 & 0.661 & 0.675 & 0.637
& 0.781 & 0.656 \\

ChohoTech 
& 0.781 & 0.780 & 0.765 & 0.761
& 0.672 & 0.625 & 0.627 & 0.586
& 0.772 & 0.628 \\

YY-LAB 
& 0.705 & 0.748 & 0.684 & 0.726
& 0.576 & 0.601 & 0.719 & 0.569
& 0.716 & 0.616 \\

YN-LAB 
& 0.657 & 0.610 & 0.656 & 0.667
& 0.539 & 0.511 & 0.538 & 0.522
& 0.648 & 0.528 \\

IGIP-LAB 
& 0.523 & 0.634 & 0.636 & 0.590
& 0.410 & 0.505 & 0.519 & 0.445
& 0.596 & 0.470 \\

3DIMLAND 
& 0.575 & 0.551 & 0.594 & 0.621
& 0.459 & 0.459 & 0.457 & 0.459
& 0.585 & 0.459 \\
         
\bottomrule
\end{tabular}
}
\caption{Overall and per-category mAP and mAR scores for all participating teams in the 3DTeethLand challenge.}
\label{tab:challenge2_result}
\end{table*}

%%%
%%%

%%%
%%%%%
%%%
\clearpage
\section{Code availability}

To advance research in this field and encourage widespread adoption within the community, the dataset is integrated within PyTorch Geometric \\ \textit{torch\_geometric.datasets.Teeth3DS} \footnote{\url{https://pytorch-geometric.readthedocs.io/en/latest/generated/torch_geometric.datasets.Teeth3DS.html}}, a widely adopted standard framework for deep learning research.  This integration allows researchers to directly load the scans, labels, and metadata.  The evaluation code is also publicly available 
\footnote{\url{https://github.com/abenhamadou/3DTeethSeg_MICCAI_Challenges}}
and 
\footnote{\url{https://github.com/crns-smartvision/3DTeethLand}}.

\section{Discussion}
The introduction of the first publicly available dataset for 3D intraoral scan analysis provides a crucial resource to advance research and foster community engagement in this clinically significant area. Despite its value, fully automated CAD systems still require substantial improvements across all core tasks. For instance, depending on application requirements, tooth-missing detection should ideally be below 1/100, highlighting the need for future refinements in tasks and evaluation metrics to assess algorithm performance under such conditions.

Beyond the current tasks, the dataset should support applications such as 3D tooth modeling, reconstruction, and detection of dentition anomalies. 

Enriching future versions of the dataset is essential to increase its representativeness of diverse clinical scenarios, including broader demographic coverage, a wider range of dental conditions, and more challenging real-world situations, such as noisy or incomplete scans. To this end, we are currently undertaking new data collection and annotation sessions, aiming to continuously improve the dataset and release updated versions. These enhancements would further enable the development of robust and generalizable models not only for landmark identification and segmentation but also for other emerging tasks.

\section{Conclusion}

In this paper, we present Teeth3DS+, an extended benchmark for intraoral 3D scan analysis. This dataset is the first publicly available resource specifically designed to support MICCAI challenges, enabling comprehensive 3D intraoral scan analysis across multiple tasks, including tooth detection, segmentation, labeling, and landmark identification. The dataset has been carefully validated both technically and clinically, ensuring high-quality annotations and a reliable benchmark. Additionally, a standardized evaluation protocol is provided to enable fair and consistent comparison of different methods on the dataset.

\section*{Acknowledgments}

The authors would like to thank Udini for helping with clinical data collection, as well as the clinical experts for their thorough validation of the annotated data.

\bibliographystyle{elsarticle-num} 
\bibliography{main}
\end{document}